\documentclass[sigconf]{acmart}

\usepackage{graphicx}
\usepackage{caption}
\usepackage{subcaption}
\usepackage{listings}
\usepackage{xcolor}
\usepackage{multirow}

\definecolor{codegreen}{rgb}{0,0.6,0}
\definecolor{codegray}{rgb}{0.5,0.5,0.5}
\definecolor{codepurple}{rgb}{0.58,0,0.82}
\definecolor{backcolour}{rgb}{0.95,0.95,0.92}

\lstdefinestyle{mystyle}{
    commentstyle=\color{codegreen},
    keywordstyle=\color{magenta},
    numberstyle=\tiny\color{codegray},
    stringstyle=\color{codepurple},
    basicstyle=\ttfamily\scriptsize,
    breakatwhitespace=false,         
    breaklines=true,                 
    captionpos=b,                    
    keepspaces=true,                 
    numbers=left,                    
    numbersep=5pt,                  
    showspaces=false,                
    showstringspaces=false,
    showtabs=false,                  
    tabsize=2
}
\AtBeginDocument{%
  \providecommand\BibTeX{{%
    \normalfont B\kern-0.5em{\scshape i\kern-0.25em b}\kern-0.8em\TeX}}}

\setcopyright{acmcopyright}
\copyrightyear{2023}
\acmYear{2023}
\acmDOI{10.1145/3605573.3605613}

\acmConference[ICPP 2023]{52nd International Conference on Parallel Processing}{August 7--10, 2023}{Salt Lake City, UT, USA}
\acmBooktitle{52nd International Conference on Parallel Processing (ICPP 2023), August 7--10, 2023, Salt Lake City, UT, USA}
\acmPrice{15.00}
\acmISBN{978-1-4503-XXXX-X/18/06}

\begin{document}

\title{Colossal-AI: A Unified Deep Learning System For Large-Scale Parallel Training}

\author{Shenggui Li}
\email{lisg@hpcaitech.com}
\orcid{0000-0003-2037-2496}
\affiliation{
  \institution{HPC-AI Technology Inc.}
  \country{Singapore}
}

\author{Hongxin Liu}
\email{liuhongxin@hpcaitech.com}
\orcid{0009-0001-0495-1108}
\affiliation{
  \institution{HPC-AI Technology Inc.}
  \country{China}
}

\author{Zhengda Bian}
\email{bian.zhengda@hpcaitech.com}
\orcid{0000-0002-1906-1781}
\affiliation{
  \institution{HPC-AI Technology Inc.}
  \country{China}
}

\author{Jiarui Fang}
\email{fangjiarui123@gmail.com}
\orcid{0000-0002-6724-2763}
\affiliation{
  \institution{HPC-AI Technology Inc.}
  \country{China}
}

\author{Haichen Huang}
\email{haichen.cs@hotmail.com}
\orcid{0009-0004-6301-5557}
\affiliation{
  \institution{HPC-AI Technology Inc.}
  \country{China}
}

\author{Yuliang Liu}
\email{yuliangliu888@gmail.com}
\orcid{0009-0003-1878-0199}
\affiliation{
  \institution{HPC-AI Technology Inc.}
  \country{China}
}

\author{Boxiang Wang}
\email{bwang@g.harvard.edu}
\orcid{0000-0003-3622-6020}
\affiliation{
  \institution{HPC-AI Technology Inc.}
  \country{Singapore}
}

\author{Yang You}
\email{youy@comp.nus.edu.sg}
\orcid{0000-0003-2816-4384}
\affiliation{
  \institution{National University of Singapore}
  \country{Singapore}
}

\renewcommand{\shortauthors}{Li, et al.}

\begin{abstract}
The success of Transformer models has pushed the deep learning model scale to billions of parameters, but the memory limitation of a single GPU has led to an urgent need for training on multi-GPU clusters.
However, the best practice for choosing the optimal parallel strategy is still lacking, as it requires domain expertise in both deep learning and parallel computing.
The Colossal-AI system addressed the above challenge by introducing a unified interface to
scale your sequential code of model training to distributed environments.
It supports parallel training methods such as data, pipeline, tensor, and sequence parallelism and is integrated with heterogeneous training and zero redundancy optimizer. 
Compared to the baseline system, Colossal-AI can achieve up to 2.76 times training speedup on large-scale models.

\end{abstract}

\ccsdesc[500]{Computing methodologies~Machine learning algorithms}
\ccsdesc[500]{Computing methodologies~Parallel computing methodologies}

\keywords{datasets, neural networks, gaze detection, text tagging}



\maketitle


\section{Introduction}

Deep learning has been successful in many applications and brought breakthroughs in difficult problems. With large amounts of data, neural networks like BERT~\cite{devlin2018bert} and Vision Transformer~\cite{dosovitskiy2020image} are capable of learning high-dimensional features and making predictions on a level even humans cannot match. As powerful hardware becomes available, neural networks have more diverse architectures and a larger number of parameters. The AI community has seen a trend of deep learning models becoming larger, with an array of large-scale models ranging from BERT-Large, GPT-2~\cite{Radford2019LanguageMA} (1.5 billion parameters), GPT-3~\cite{brown2020language} (175 billion parameters), to GLM~\cite{du2021all} (1.75 trillion parameters). These large-scale models require more data and computing resources but also have better generality and performance. As more robust computing hardware and larger datasets become available, the trend is expected to continue and traditional training methods will become less effective, making distributed training necessary for large-scale model training.

The limited fast memory of commonly used accelerator hardware, such as GPU, is a bottleneck of scaling the model to billions of parameters. 
The memory consumption in deep learning comes from model parameters, layer activations, gradients, and optimizer states. 
We refer to model parameters, gradients, and optimizer states as model data and layer activations as non-model data. 
When training with adaptive optimizers~\cite{duchi2011adaptive, kingma2014adam}, the total memory consumption of model data can be several times larger than that consumed by parameters alone, making a single GPU no longer sufficient for large-scale model training.
10 billion parameters in FP16 format can already consume 20 GB of model memory, while a typical GPU only has 16 or 32 GB of memory. Without any optimization, training a model of 10 billion parameters with one data sample can cost more than 80 GB of memory, which is far more than that of a typical GPU.

Data parallelism~\cite{horovod} has scaled models such as ResNet~\cite{resnet} to multi-GPU training and other methods such as activation checkpointing~\cite{act_ckpt} were proposed to reduce the non-model data by trading computation for memory. However, these methods failed to cope with billion-parameter model data. Parallelization techniques such as pipeline parallelism~\cite{NEURIPS2019_093f65e0, 10.1145/3341301.3359646} and tensor parallelism~\cite{shoeybi2019megatron} were explored to shard the model data, making it possible to train models at a larger scale. 
The current state-of-the-art systems which provide a solution to the scaling challenge include GShard~\cite{lepikhin2021gshard}, FairScale~\cite{FairScale2021}, Megatron-LM~\cite{megatron} and DeepSpeed~\cite{rasley2020deepspeed}. Among these systems, Megatron-Lm and DeepSpeed are the most popular in the open-source community and deliver the best performance. Thus, they are chosen as the baseline of our experiments. Megatron-LM trains Transformer-based models by utilizing optimized pipeline and tensor parallelism. 
Meanwhile, DeepSpeed proposed an efficient method to partition the model-related data to fully eliminate memory redundancy in data parallel training. 
These two efficient methods paved the way to scale model training to hundreds of devices and billions of parameters. 

\begin{figure}[h]
    \centering
    \includegraphics[width=0.25\textwidth]{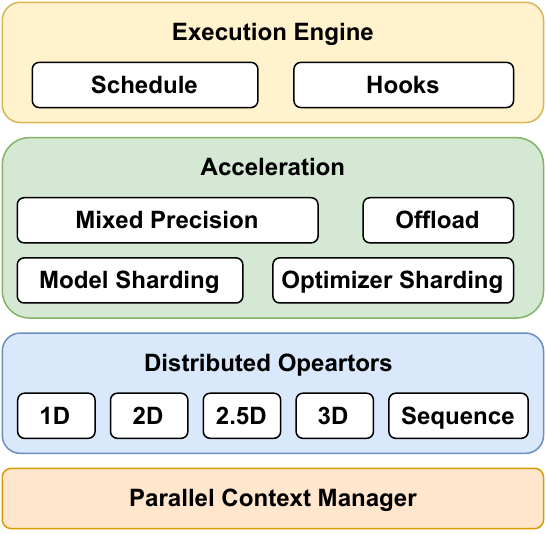}
    \caption{Architecture of Colossal-AI}
    \label{fig: arch}
\end{figure}

As most deep learning engineers and researchers are used to writing non-distributed code, it is reasonably difficult for them to adapt to parallel and distributed programming. The existing systems either introduce extra complexity in parallelizing the model training or offer insufficient parallelization techniques. We have thus developed Colossal-AI, which is an open-source system to democratize complicated distributed training in the AI community by unifying an array of training acceleration techniques in one deep learning system. In this system, we also included novel parallelism methods such as multi-dimension tensor parallelism and sequence parallelism. Colossal-AI aims to make distributed training easy by providing user-friendly APIs while allowing users to maintain their coding habit of writing single-node programs. In a nutshell, we bring the following major contributions to large-scale distributed training in this work:

\begin{itemize}
\item Colossal-AI is a unified deep learning system that provides the fullest set of acceleration techniques for the AI community. With its modular design as shown in Figure~\ref{fig: arch}, Colossal-AI allows for free combination of these techniques to achieve the best training speedup. The details of the system architecture will be discussed in the Implementation section.
\item Optimized parallelism and heterogeneous training methods are provided in Colossal-AI. These methods achieve better system performance than the baseline systems. They are provided for the user via friendly APIs with minimum code changes.
\item In-depth analysis was conducted to investigate the suitable parallelism strategies under different hardware conditions.
\end{itemize}

\section{Background}

\begin{figure}[h]
    \centering
    \includegraphics[width=0.4\textwidth]{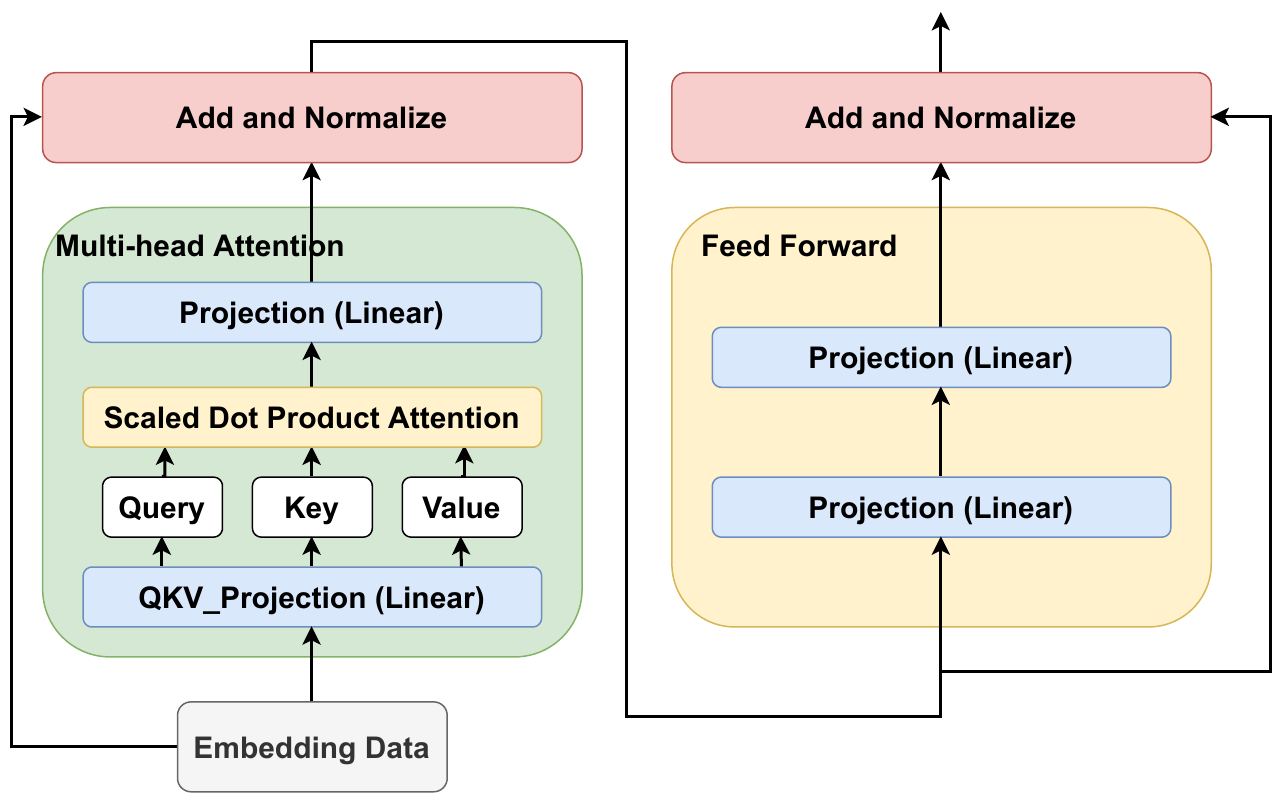}
    \caption{Architecture of The Transformer Layer}
    \label{fig: transformer-layer}
\end{figure}

Thanks to the advent of the Transformer architecture, deep learning models have gained unprecedented performance improvement in domains such as Computer Vision and Natural Language Processing. The typical architecture of a Transformer layer consists of a Multi-head Attention block and a Feed Forward block as shown in Figure~\ref{fig: transformer-layer}. This architecture can scale to billions of parameters and larger models can deliver more impressive performance improvement. For example, GPT-3~\cite{brown2020language} outperforms the smaller models by $18\%$ absolute increase in prediction accuracy on the LAMBADA language task~\cite{lambada}. 

To cope with the increasing model size, AI engineers have explored distributed training in pursuit of lower time costs.
Various techniques were proposed to accelerate distributed training and they will be discussed below.

\begin{figure*}[h!]
    \begin{subfigure}[h!]{0.3\textwidth}
        \centering
        \includegraphics[height=3cm]{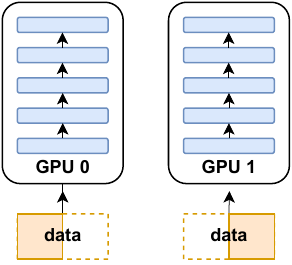}
        \caption{data parallel}
        \label{fig: data-parallel}
    \end{subfigure}%
    ~ 
    \begin{subfigure}[h!]{0.3\textwidth}
        \centering
        \includegraphics[height=3cm]{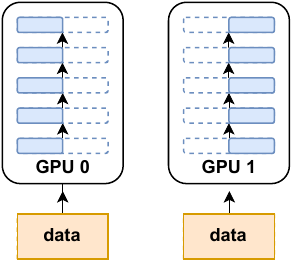}
        \caption{tensor parallel}
        \label{fig: tensor-parallel}
    \end{subfigure}
    ~
    \begin{subfigure}[h!]{0.3\textwidth}
        \centering
        \includegraphics[height=3cm]{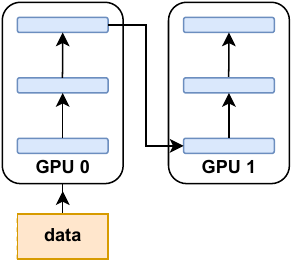}
        \caption{pipeline parallel}
        \label{fig: pipeline-parallel}
    \end{subfigure}
    
    \caption{Existing parallelism for distributed training}
\end{figure*}

\subsection{Data Parallelism}

Data parallelism is the most common parallelism technique due to its simplicity. In data parallel training, the model is replicated across the devices, and the dataset is split into several shards. Each dataset shard is fed to the model on one device as shown in Figure~\ref{fig: data-parallel}. Collective communication is required to synchronize the parameter gradients after backward propagation ~\cite{10.14778/3415478.3415530}. Data parallelism makes it easy to train a model on multiple devices and scales sub-linearly with the number of devices.

One problem of data parallelism is that each device holds a copy of the model parameters, optimizer states, and gradients, leading to memory redundancy. 
When using stateful optimizers such as Adam~\cite{adamoptim}, the optimizer states (i.e. momentum and variance) can occupy three times larger memory space compared to that occupied by the model parameters ~\cite{adamoptim, https://doi.org/10.48550/arxiv.2108.05818}. 
To eliminate such redundancy, Zero Redundancy Optimizer was proposed in DeepSpeed~\cite{rasley2020deepspeed} to partition these redundant model data over different devices during data parallel training. 
As each device only holds a partition of gradients, optimizer states, and parameters, it will only update the partitioned parameters instead of the full model parameters on one device.

\subsection{Model Parallelism}

To go beyond data parallel training, more techniques were explored to shard the model parameters over a larger number of devices. As a result, model parallelism was proposed to tackle this problem. There are generally two types of model parallelism: tensor parallelism and pipeline parallelism.

1) Tensor Parallelism

Tensor parallelism shards the tensor over an array of devices and requires a distributed matrix-matrix multiplication algorithm for arithmetic computation as shown in Figure~\ref{fig: tensor-parallel}. Megatron-LM~\cite{shoeybi2019megatron} proposed 1D tensor parallelism which splits the linear layer in the row or column dimensions for the Transformer architecture~\cite{attnisalluneed}. More advanced tensor sharding mechanisms such as 2D~\cite{xu2021_2d}, 2.5D~\cite{wang_2p5d}, and 3D~\cite{bian2021_3d} were proposed to shard tensors in more dimensions. Collective communication is required among devices to ensure arithmetic correctness. 

\begin{figure}[h]
    \centering
    \includegraphics[width=0.4\textwidth]{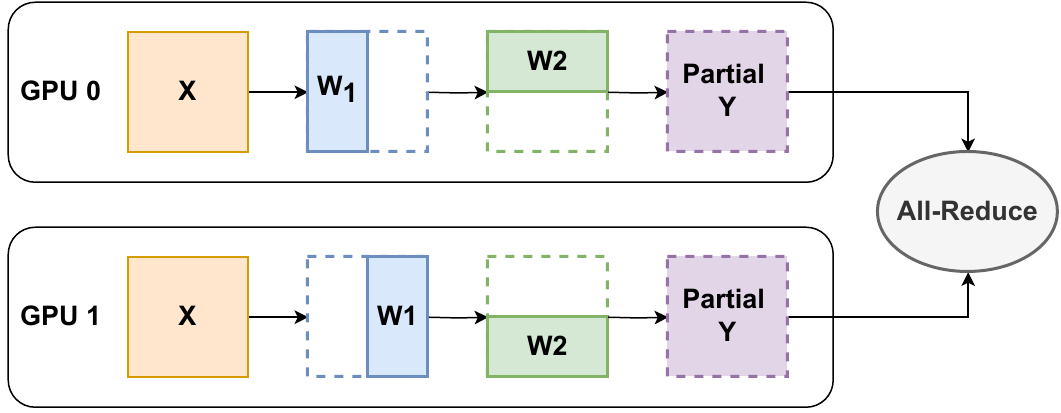}
    \caption{Megatron-LM MLP Module}
    \label{fig: megatron-mlp}
\end{figure}

In Meagtron-LM, the tensors are sharded in one dimension. Taking the Feed Forward module of the Transformer layer as an example, we can view the module as a matrix multiplication of $Y = W_2W_1X$ as shown in Figure~\ref{fig: megatron-mlp}, where $X$ is the input, $W_1$ and $W_2$ are the model parameters. $W_1$ and $W_2$ can be sharded vertically and horizontally respectively and produce a partial result of $Y$ on each device. An all-reduce operation can be applied to the partial result to obtain the correct final result of the matrix multiplication. In this way, each device will only hold $1/N$ of the parameters when training on $N$ devices. This allows the model size to scale beyond the memory capacity of a single device.

One of the major problems of the 1D method is that it assumes the interconnect of devices has the same bandwidth. This makes it friendly only on machines with fully connected NVLinks among the GPUs on a single node as shown in Figure~\ref{fig:full-nvlinks}. However, such high-end hardware is expensive and scarce. Many machines, even some in the supercomputing centers, only have partially connected GPUs as shown in Figure~\ref{fig:pair-nvlinks}. With this kind of GPU topology, the communication bandwidth between distant devices via the PCIe bus is much lower than that of directly connected GPUs. Therefore, the low communication bandwidth can hinder the efficiency of all-reduce operations in 1D tensor parallelism.

In addition, the 1D tensor parallelism has redundant memory usage in layer inputs and outputs. Taking the Feed Forward layer in the Transformer architecture as an example, the input $X$ and output $Y$ of the MLP layer are duplicated across different devices as shown in Figure~\ref{fig: megatron-mlp}. Such memory redundancy limits the maximum model size which can be trained on limited hardware resources, and is not helpful with the democratization of large-scale distributed training.

Besides 1D tensor parallelism, more advanced tensor parallelism is introduced for large-scale model training, namely 2D, 2.5D, and 3D tensor parallelism ~\cite{xu2021_2d, wang_2p5d, bian2021_3d}. These methods split input, weight, and output tensors and thus have advantages in memory and communication efficiency, better coping with different hardware specifications. This provides the user with an option of using the most suitable tensor parallelism technique for their machines.

2D tensor parallelism~\cite{xu2021_2d} relies on the SUMMA and Cannon matrix multiplication algorithm~\cite{cannon1969cellular,berntsen1989communication,10.5555/899248} and splits a tensor along two different dimensions. Given $N$ devices arranged in a square network topology, a tensor of shape $[P, Q]$ will be partitioned into a chunk tensor of shape $[P/\sqrt{N}, Q/\sqrt{N}]$. 2.5D Tesnor Paralleism~\cite{wang_2p5d} was inspired by 2.5D matrix multiplication algorithm~\cite{Solomonik2011CommunicationOptimalP2} and proposed to further parallelize 2D tensor parallelism. It adds the optional $depth$ dimension of the matrix for parallelization. When $depth=1$, it is close to 2D tensor parallelism. When $depth > 1$, it partitions the matrix 3 times and adds one more degree of parallelization. Given $N$ devices, the tensor is split in a way such as $N = S^2*D$ where $S$ is the size of one side of the square and $D$ is the depth of the cuboid. 3D tensor parallelism~\cite{bian2021_3d} was proposed based on the 3D matrix multiplication algorithm~\cite{5389455}. 3D tensor parallelism splits a tensor in a cubic manner. As not every tensor has 3 dimensions, we choose to partition the first and last dimension only where the first dimension will be partitioned twice. For example, a tensor of shape $[P, Q]$ will be partitioned into a chunk tensor of shape $[P/\sqrt[3]{N}^2, Q/\sqrt[3]{N}]$.

As the advanced tensor parallelism methods require different network topologies, the user needs to choose the method based on the number of GPUs. 1D method can work with any number of GPUs while 2D, 2.5D and 3D methods require the $n^2$, $a*n^2$, and $n^3$ GPUs respectively, where $a$ and $n$ are positive integers. The user can fall back to 1D tensor parallelism when the number of GPUs does not fulfill the requirement. These advanced tensor parallelism methods provide lower communication volume when scaling to a larger number of devices \cite{cannon1969cellular, berntsen1989communication, Solomonik2011CommunicationOptimalP2, 5389455} and this will be further discussed in Section 3.1. 

2) Pipeline Parallelism

Methods such as PipeDream~\cite{pipedream}, GPipe~\cite{gpipe}, and Chimera~\cite{chimera} were proposed to split the model into several chunks of consecutive layers and each chunk is allocated to a device as shown in Figure~\ref{fig: pipeline-parallel}. Intermediate activations and gradients are passed between pipeline stages to complete the forward and backward pass. As a result, this method reduces cross-node communication. Pipeline parallelism allows multiple devices to compute simultaneously, leading to a higher throughput. One drawback of pipeline parallel training is that there will be some bubble time, where some devices are idle when others are engaged in computation, leading to the waste of computational resources ~\cite{pipedream}.

\subsection{Sequence Parallelism}

Tensor parallelism mainly tackles the memory bottleneck brought by model data. However, the non-model data can be the bottleneck in applications such as AlphaFold and document-level text understanding. This is because these applications rely on long-sequence data. As the self-attention module in the Transformer layer is of quadratic complexity with respect to the sequence length, long-sequence data will increase the memory usage consumed by the intermediate activation, limiting the training capability of the devices. 

Sequence parallelism~\cite{seq_parallel} is proposed to enable long-sequence modeling by breaking the memory wall brought by the large sequence dimension. In sequence parallelism, the model is replicated across devices just like data parallelism. The input data is split along the sequence dimension and each device only keeps a sub-sequence. The self-attention module is replaced with the Ring Self-Attention module such that the partial query, key, and value embeddings are exchanged among devices to complete the self-attention calculation.

\subsection{Heterogeneous Training}

To further expand the memory capacity of a single device, DeepSpeed proposed zero-offload~\cite{ren2021zerooffload} which moves the tensors from GPU to CPU or NVMe disks when not in use to make room for larger models. It is often seen that CPU memory is much larger than the GPU memory on machines such as the Nvidia DGX1 workstation. By utilizing high-performance heterogeneous storage devices and appropriately swapping tensors between different hardware devices, it became possible to train a model with billions of parameters on a single GPU. This is especially friendly to users with limited computing resources and essential for the democratization of large-scale model training.

\subsection{Automatic Parallelization}

The latest advance in parallel training is the automatic selection and execution of parallelization strategies as demonstrated in FlexFlow~\cite{flexflow} and Alpa~\cite{alpa}. Alpa was proposed recently to automatically search for a suitable parallelization plan including data and model parallelism given the cluster mesh. It then compiles the computation graph into distributed sharded graph with communication operators and runs the compiled executable on the cluster. However, Alpa is not made to be hardware-aware and does not automatically consider the network topology. Meanwhile, it does not search for other optimization techniques such as activation checkpointing, leading to suboptimal results.


\section{Design}

Colossal-AI is featured by an array of acceleration techniques constructed in a modular way, which can cover a wide range of training settings to achieve maximal performance. It addresses the difficulties in achieving consistent acceleration in deep learning training due to diverse hardware conditions, met by Megatron-LM and DeepSpeed as well. This section will discuss the implementation and analysis of the acceleration techniques integrated in Colossal-AI.

\subsection{Multi-dimensional model parallelism}

First of all, Colossal-AI provides an array of  model parallelism methods to cater to the needs of distributed training. Thus, it allows the model size to scale to billions of parameters by sharding the model over devices. In Colossal-AI, all existing tensor parallelism methods are supported so that the user can choose one method based on their training requirements and the number of GPUs while Megatron-LM only supports 1D tensor splitting. As tensor parallelism is mainly applied to matrix-matrix multiplication, it is highly suitable for the acceleration of Transformer models which widely uses linear layers.

Among all tensor parallelism methods, one prominent advantage of advanced tensor parallelism, namely 2D, 2.5D, and 3D tensor parallelism, is that it has a lower communication cost compared to 1D tensor parallelism. Table~\ref{tab: tp-comm-vol} has shown the total communication volume when computing a matrix multiplication $Y=WX$ where $X$ is of shape $(b, s, h)$, $W$ is of shape $(h, h)$ and $Y$ is of shape $(b, s, h)$. In Table~\ref{tab: tp-comm-vol}, the following notations are used.

\begin{itemize}
  \item $p$: the total number of GPUs
  \item $j$: the number of GPUs on one side of the square-shaped network topology, where $p = j^2$
  \item $k$: the number of GPUs on one side of the front square of the cuboid-shaped network topology, where $p = d*k^2$
  \item $d$: the number of GPUs in the depth dimension of the cuboid-shaped network topology
  \item $l$: the number of GPUs on one side of the cube-shaped network topology, where $p = l^3$
  \item $S_{x}$: the number of elements in the input matrix $X$, the same semantic is applied to $S_{W}$ and $S_{Y}$.
  \item $b$: batch size of the input matrix $X$.
  \item $s$: the sequence length of the input matrix $X$.
  \item $h$: the hidden size of the weight $W$.
\end{itemize}

\begin{table}[H]
\centering
\begin{tabular}{|p{1.2cm}|p{5cm}|} 
 \hline
 Mode & Total Communication Volume (number of elements transferred) \\ [0.5ex] 
 \hline\hline
 1D & $2(p-1)*S_{x}$ \\ 
 \hline
 2D & $3(j-1)*(S_{x}+S_{w})$ \\ 
 \hline
 2.5D & $3(k-1) * (S_{x}/d + S_{w})$  \\ 
 \hline
 3D & $2(l-1)/l * (S_{x} + S_{w} + S_{y})$ \\ 
 \hline
\end{tabular}
\caption{Communication Volume of Tensor Parallelism}
\label{tab: tp-comm-vol}
\end{table}

\begin{figure}[H]
    \centering
    \includegraphics[width=0.3\textwidth]{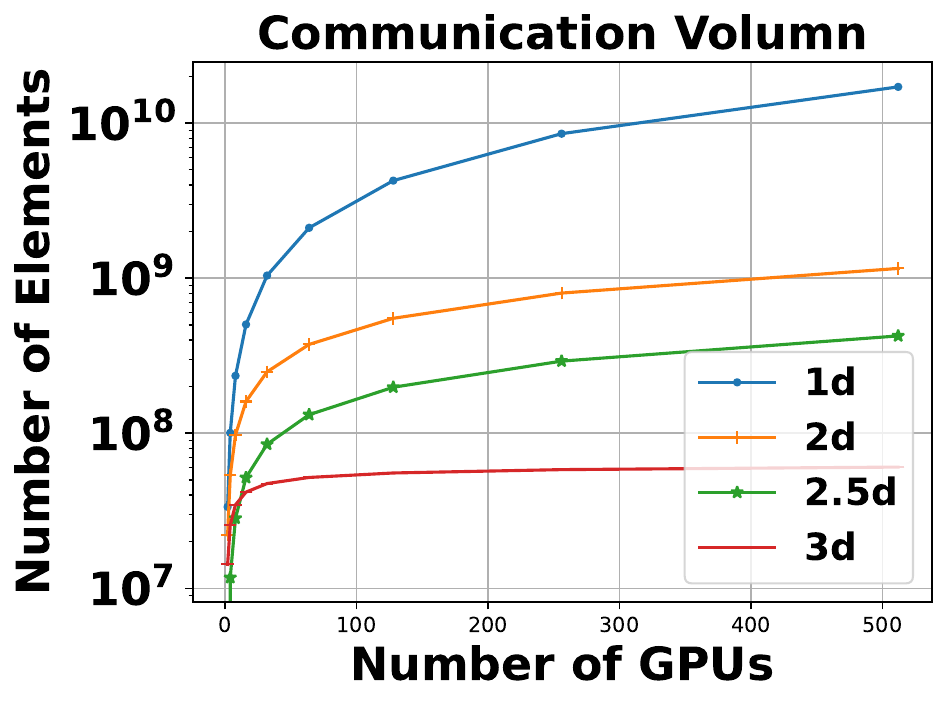}
    \caption{Scaling Performance of Tensor Parallelism in Theoretic Analysis ($h=1024, s=512, b=32$)}
    \label{fig: tp-scaling-analysis}
\end{figure}

As shown in Figure~\ref{fig: tp-scaling-analysis}, the communication volume of the advanced tensor parallelism is significantly lower than that of 1D tensor parallelism, especially when a large number of nodes is used. The underlying reason for communication efficiency is that advanced tensor parallelism only incurs communication on a sub-group of the computing nodes. For example, in 2D parallelism, collective communication only involves the nodes in one row or one column of the square-shaped network. In contrast, 1D tensor parallelism involves all computing nodes for one collective communication call. Therefore, advanced tensor parallelism allows scaling beyond one node while 1D tensor parallelism is often restricted to intra-node computing.

Besides tensor parallelism, Colossal-AI has also included sequence parallelism and pipeline parallelism so that hybrid parallelism is available out of the box to accelerate model training in large-scale clusters. 

\subsection{Enhanced Sharding and Offloading}\label{sec:sharding&offloading}
\label{zero-optim}

Zero redundancy data parallel training and offloading proposed by DeepSpeed enable large-scale model training. However, it is still bound to the CPU-GPU and GPU-GPU communication and its rigid implementation leads to poor extensibility.
Colossal-AI has re-designed the tensor sharding and offloading mechanism for better performance. Colossal-AI proposed a unified sharded tensor interface and supports customizable sharding strategies and life-cycle hooks for easy modification of the training workflow. As such, zero-redundancy data parallel can be easily supported and extended. Meanwhile, it also integrates the chunk strategy proposed in PatrickStar~\cite{https://doi.org/10.48550/arxiv.2108.05818} to arrange tensors in chunks to further improve the communication bandwidth utilization and memory usage, making tensor offloading more efficient.

\begin{figure}[h]
    \centering
    \includegraphics[width=0.4\textwidth]{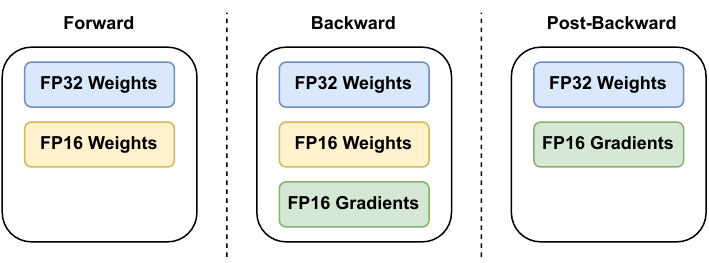}
    \caption{Memory Space Reuse}
    \label{fig: reuse}
\end{figure}

Such flexible design brings several benefits. Firstly, it enables the re-use of FP16 storage space in the memory so that larger models can be trained. In the forward pass, we hold FP16 parameters. In the backward pass, when the gradients are computed, the FP16 parameters are no longer needed. We can thus save FP16 gradients in the same storage space which holds FP16 parameters during forward as shown in Figure~\ref{fig: reuse}. In this way, Colossal-AI further reduces redundancy and peak memory usage and the CPU memory can afford to accommodate larger models.

Secondly, an adaptive tensor placement and parameter update can be enabled during heterogeneous training. In DeepSpeed's zero offloading, it provides an implementation of CPU Adam to update the model parameters in the CPU. However, this method requires all the FP32 master model weights to be placed in the CPU memory. In Colossal-AI, we implemented an adaptive hybrid Adam optimizer instead. During heterogeneous training, Colossal-AI's hybrid Adam optimizer monitors the available memory space on the GPU. It does not statically keep all FP32 weights in the CPU memory, instead, it dynamically keeps part of parameters and gradients on the GPU as long as there is space left. In this way, parameters are updated on both CPU and GPU, leading to better resource utilization and lower communication cost.

\subsection{Automatic Parallelization on Dynamic Computation Graph}\label{sec: automatic-parallelization}

Inspired by Alpa~\cite{alpa}, Colossal-AI has included an experimental automatic parallelism feature to improve upon the Alpa project. One challenge in automatic parallelization is the sharded tensor conversion. For example, a tensor sharded on its 0th dimension can be converted to the one sharded in the last dimension. Alpa hardcodes a conversion table, but this limits the number of sharded dimensions to keep the table reasonably small. We implemented a greedy algorithm to search to speed up sharding conversion and increase the number of sharding dimensions. Moreover, we integrate activation checkpoint into the search problem such that a model can be both sharded and activation checkpointed to achieve maximum performance. As this feature is only experimental, it will be discussed separately in another paper as a future work.


\section{Implementation}

The overall architecture of Colossal-AI is shown in Figure~\ref{fig: arch}. It has a parallel context manager that manages the meta information of the complex hybrid parallel distributed environment and automatically switches to the corresponding parallel mode based on the parallel context. It has a user-friendly interface for building tensor-parallel models and various acceleration tools, including activation checkpointing and mixed precision training. It also has an execution engine and trainer that provide extensibility for user customization, allowing them to define their own training schedule and hooks at the operator or trainer level.

\subsubsection{Modularity}

The principle of modularity and extensibility is upheld throughout the development and the different acceleration techniques can easily be combined in pursuit of maximal performance.

\subsubsection{Extensibility}

As a system under constant development, Colossal-AI provides various interfaces to implement customized functions for future extensions. For example, the sharding module allows the user to define their own sharding strategy and life-cycle hooks in order in an attempt to explore more efficient training methods.

\subsubsection{User-Friendliness}

To minimize the change to the user code, Colossal-AI provides user-friendly APIs for model training. The user only needs to prepare a configuration that specifies the features by following a pre-defined schema. Colossal-AI will then inject the acceleration features into the execution engine with `colossalai.initialize` as shown in Listing~\ref{code: config}.

Colossal-AI also provides parallelized popular model components such as BERT~\cite{devlin2018bert}, GPT~\cite{Radford2019LanguageMA}, ViT~\cite{dosovitskiy2020image}, which the users can use directly. This does not require the users to have domain expertise so that they do not have to manually design their parallelism strategy like GShard~\cite{lepikhin2021gshard}.

\begin{lstlisting}[language=Python,caption=Colossal-AI Usage, label={code: colossalai}]
import colossalai

# specify using 1D tensor parallelism with parallel size 4
config = dict(paralle=dict(
                tensor=dict(
                    size=4, 
                    mode='1d'
                )
            )
        )
        
# launch distributed network
colossalai.launch_from_torch(config=config)

# define your training components
...
    
# initialize with Colossal-AI
engine, trainloader, _ = \
	colossalai.initialize(model, 
	                      optimizer, 
	                      criterion, 
	                      trainloader)

# run training
for data, label in train_dataloader:
    engine.zero_grad()
    output = engine(data)
    train_loss = engine.criterion(output, label)
    engine.backward(train_loss)
    engine.step()
\end{lstlisting}


\begin{figure}[H]
    \centering
    \includegraphics[width=0.3\textwidth]{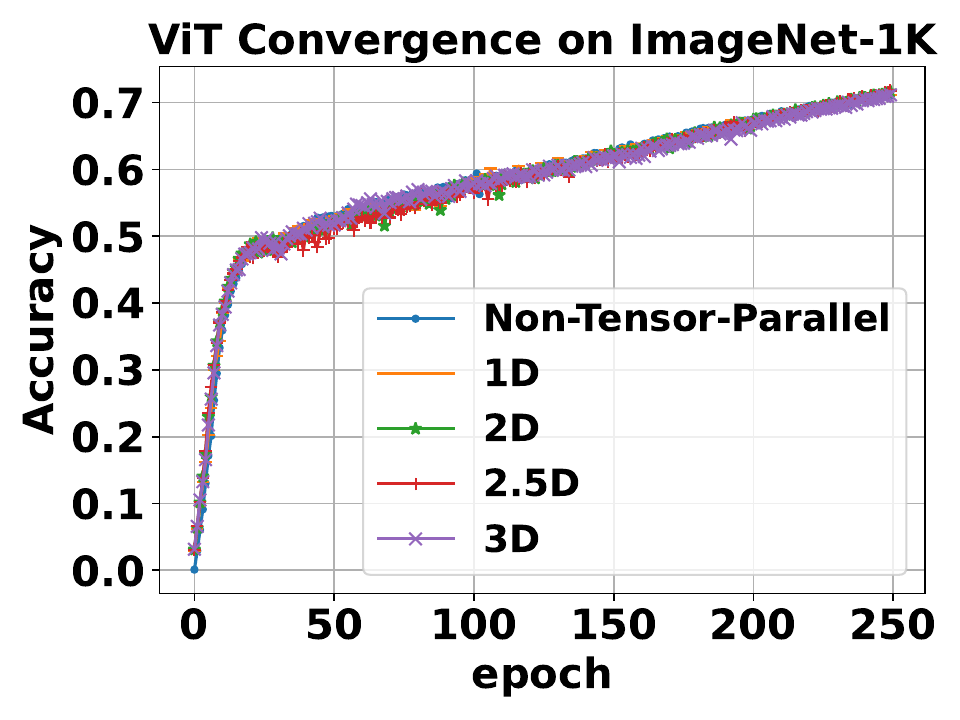}
    \caption{Convergence Performance of ViT on ImageNet}
    \label{fig: vit-convergence}
\end{figure}

\begin{table*}[t]
  \centering
  \small
  \begin{tabular}{|p{1.2cm}|p{1cm}|p{1cm}|p{2.8cm}|p{3cm}|p{3.3cm}|p{3.5cm}|}
        \hline
        System ID & \#GPUs per node & \#Nodes & GPU Model & GPU Interconnect & Cross-node Interconnect & Experiment Item \\
        \hline
        I & 8 & 1 & Nvidia A100 (80GB) & NVlink & N/A & Tensor Parallelism \\
        \hline
        II & 8 & 1 & Nvidia A100 (80GB) & NVlink between adjacent GPUs, PCIe between distant GPUs  & N/A & Tensor Parallelism, ZeRO \\
        \hline
        III & 4 & 16 & Nvidia A100 (40GB) & NVLink  & InfiniBand HDR (200Gb/s), and Dragonfly network topology & Tensor Parallelism, Sequence Parallelism \\
        \hline
        IV & 1 & 64 & Nvidia P100 (16GB) & RDMA  & Cray Aries routing and communications ASIC, and Dragonfly network topology & Tensor Parallelism \\
        \hline
  \end{tabular}
  \caption{System Specification for Experiments}
  \label{tab: sys-spec}
\end{table*}

\section{Evaluation}

\subsection{Experiment Setup}

To holistically evaluate the system performance of Colossal-AI, we have conducted various experiments on different hardware. The system specification is listed in Table~\ref{tab: sys-spec}. Due to resource constraints, we only tested a portion of the prominent features on each system as stated in the \emph{Experiment Item} column. We used Megatron-LM and DeepSpeed as our baselines for experiments and Megatron-LM tensor parallelism is annotated as 1D tensor parallelism in the results.

\subsection{Multi-Dimensional Tensor Parallelism}

1) Convergence

Experiments were conducted with Vision Transformer (ViT)~\cite{dosovitskiy2020image} on the ImageNet-1k dataset to verify the arithmetic correctness and numerical stability of multi-dimensional tensor parallelism on System III.
The ViT model has 12 Transformer layers with 384 hidden size and 6 attention heads. We used Jax initialization and AdamW optimizer with 0.003 learning rate and 0.3 weight decay. The input image is of shape 224 and the patch size is 16. The global batch size is 4k and the model is trained for 250 epochs. As shown in Figure~\ref{fig: vit-convergence}, the testing accuracy curves of Multi-Dimensional tensor parallelism well align with that of the PyTorch data parallel training.

\begin{figure*}[h]
    \begin{subfigure}[h]{0.24\textwidth}
        \centering
        \includegraphics[width=1\textwidth]{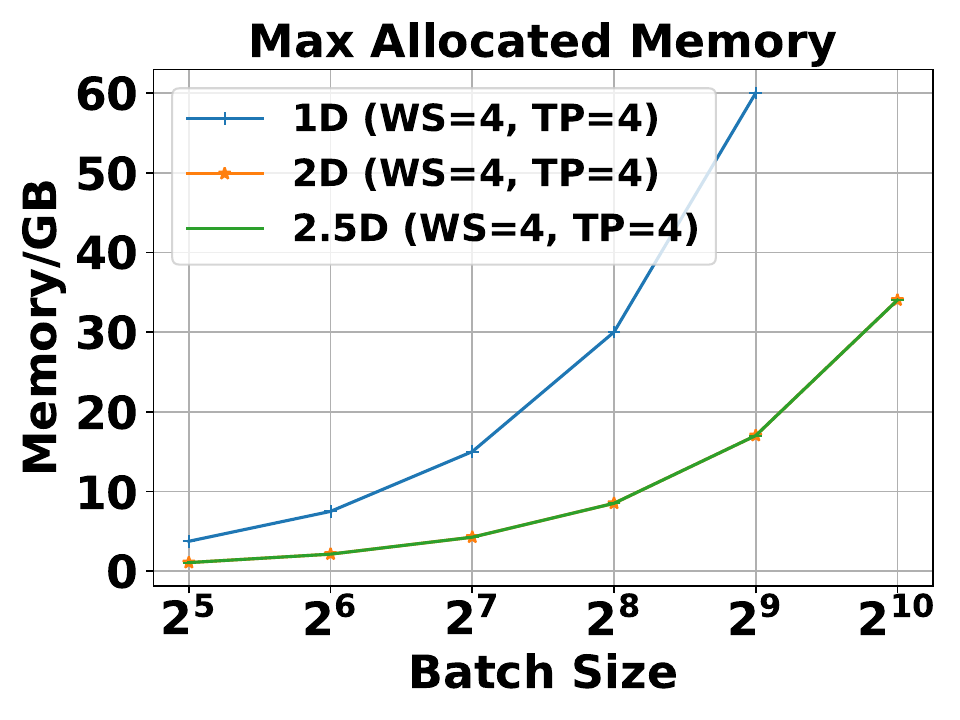}
        \caption{4 GPUs by Batch Size}
        \label{fig: 4-gpu-by-bs}
    \end{subfigure}%
    ~ 
    \begin{subfigure}[h]{0.24\textwidth}
        \centering
        \includegraphics[width=1\textwidth]{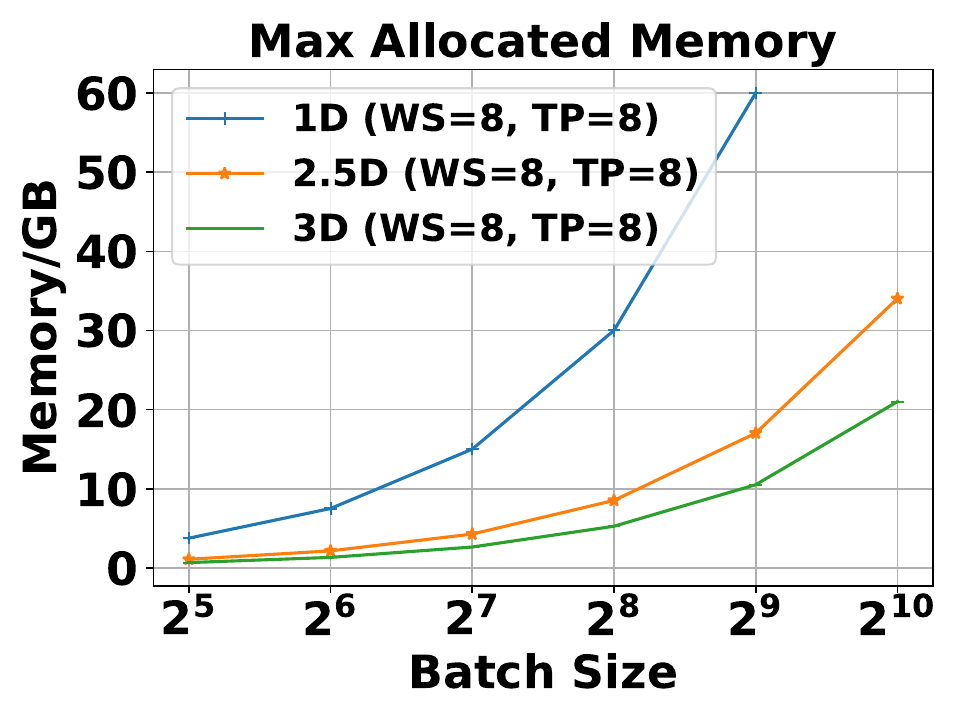}
        \caption{8 GPUs by Batch Size}
        \label{fig: 8-gpu-by-bs}
    \end{subfigure}
    ~
    \begin{subfigure}[h]{0.24\textwidth}
        \centering
        \includegraphics[width=1\textwidth]{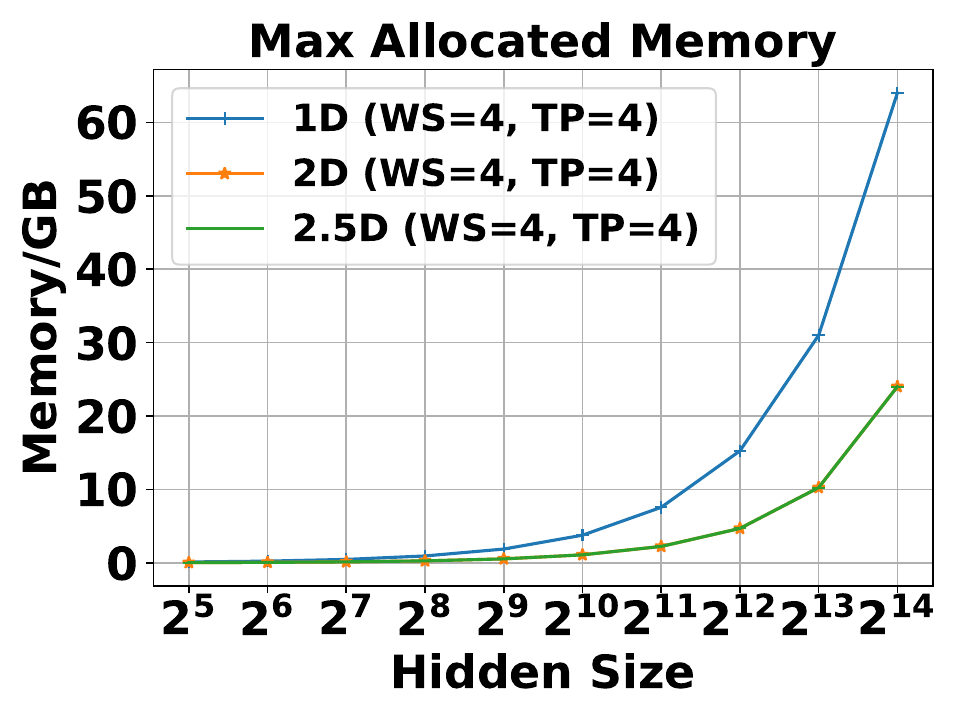}
        \caption{4 GPUs by Hidden Size}
        \label{fig: 4-gpu-by-dim}
    \end{subfigure}
    ~
    \begin{subfigure}[h]{0.24\textwidth}
        \centering
        \includegraphics[width=1\textwidth]{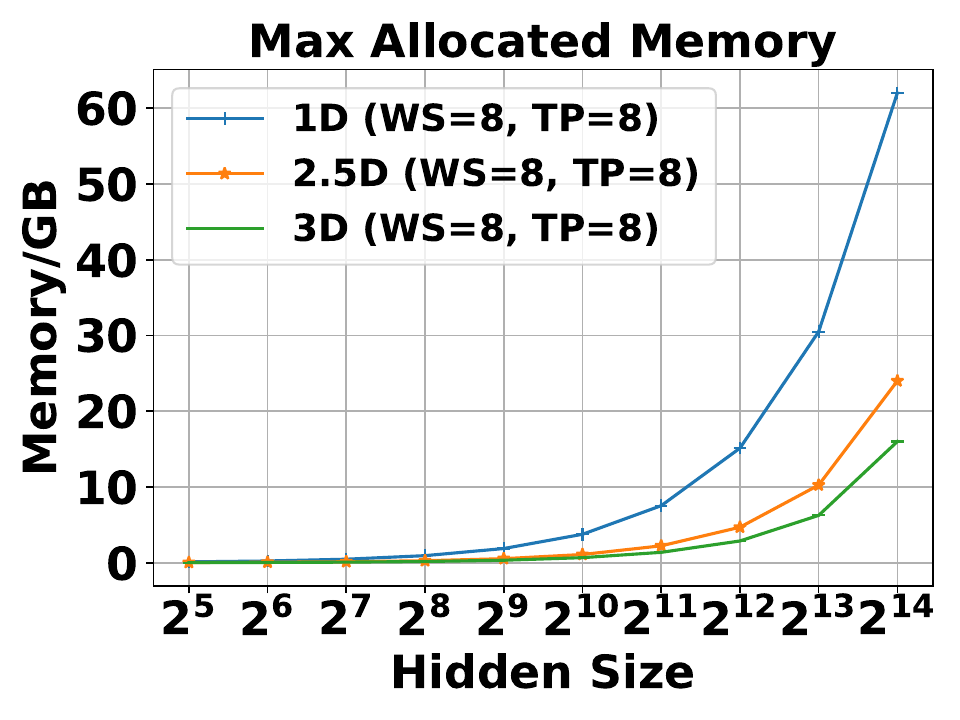}
        \caption{8 GPUs by Hidden Size}
        \label{fig: 8-gpu-by-dim}
    \end{subfigure}
    
    \caption{Range Test for Memory Consumption of Tensor Parallelism with 4/8 GPUs}
    \label{fig: range-test}
\end{figure*}

2) Memory Efficiency

As 2D, 2.5D, 3D tensor parallelisms partition the input data, layer weight, and output activation while 1D tensor parallelism only partitions the layer weight, the former is expected to have lower memory consumption. As a result, the first three methods allow the GPUs to accommodate larger models. To demonstrate the memory efficiency, we have conducted two range tests which scale by batch size and hidden size on System I. In this range test, we created a model which consists of two linear layers. We run 1D, 2D and 2.5D experiments on 4 GPUs and 1D, 2.5D (depth=2), and 3D experiments on 8 GPUs. We measure the max allocated CUDA memory during the forward and backward pass, and the results are shown in Figure~\ref{fig: range-test}. The memory consumption of 1D tensor parallelism is much higher than those of 2D, 2.5D, and 3D tensor parallelism. With the batch size equal to 512 and 8 GPUs, the memory consumption of 2.5D and 3D is 44\% and 65\% lower than that of 1D tensor parallelism respectively in Figure~\ref{fig: 8-gpu-by-bs}. With the hidden size of 16384 and 8 GPUs, the memory performance of 2.5D and 3D tensor parallelism is 62\% and 74.2\% better than that of 1D tensor parallelism respectively in Figure~\ref{fig: 8-gpu-by-dim}. Therefore, more advanced tensor parallelism is a better option when scaling to super large-scale models.

\begin{table*}[t]
    \centering
    \small
    \begin{tabular}{|p{0.7cm}|p{1.8cm}|p{2.4cm}|p{1.5cm}|p{2cm}|p{1.4cm}|p{1.5cm}|p{2cm}|} 
        \hline
        \#GPUs & Mode & \#Transformer Layer & Hidden Size & \#Attention Heads & Batch Size & Throughput (img/sec) & Speedup over 1D (\%) \\
        \hline
        \multirow{3}{*}{4}  & 1D    & \multirow{3}{*}{24} & \multirow{3}{*}{2048} & \multirow{3}{*}{32} & 128 & 5.06  & -   \\
                            & 2D    &                     &                       &                     & 256 & 6.18  & 22.1 \\
                            & 2.5D  &                     &                       &                     & 256 & 6.73  & 33.0 \\
        \hline
        \multirow{3}{*}{8}  & 1D    & \multirow{3}{*}{24} & \multirow{3}{*}{2048} & \multirow{3}{*}{32} & 256 & 7.46  & -  \\
                            & 2.5D  &                     &                       &                     & 384 & 6.57  & -11.9 \\
                            & 3D    &                     &                       &                     & 512 & 8.38  & 12.3  \\
        \hline
        \multirow{3}{*}{16} & 1D    & \multirow{3}{*}{32} & \multirow{3}{*}{4096} & \multirow{3}{*}{64} & 64  & 3.42  & -    \\
                            & 2D    &                     &                       &                     & 256 & 5.33  & 55.8 \\
                            & 2.5D  &                     &                       &                     & 256 & 5.46  & 59.6 \\
        \hline
        \multirow{3}{*}{32} & 1D    & \multirow{3}{*}{32} & \multirow{3}{*}{4096} & \multirow{3}{*}{64} & 128 & 4.22  & -    \\
                            & 2.5D  &                     &                       &                     & 256 & 5.46  & 50.6 \\
        \hline
        \multirow{4}{*}{64} & 1D    & \multirow{4}{*}{32} & \multirow{4}{*}{4096} & \multirow{4}{*}{64} & 128 & 4.63  & - \\
                            & 2D    &                     &                       &                     & 512 & 12.76 & 275.5 \\
                            & 2.5D  &                     &                       &                     & 512 & 4.93  & 6.5 \\
                            & 3D    &                     &                       &                     & 512 & 8.63  & 86.4 \\
        \hline
  \end{tabular}
  \caption{Performance of Tensor Parallelism with Different Number of GPUs}
\label{tab: perf-with-gpu}
\end{table*}

\begin{figure}[htp]
    \begin{subfigure}{0.2\textwidth}
        \centering
        \includegraphics[width=\textwidth]{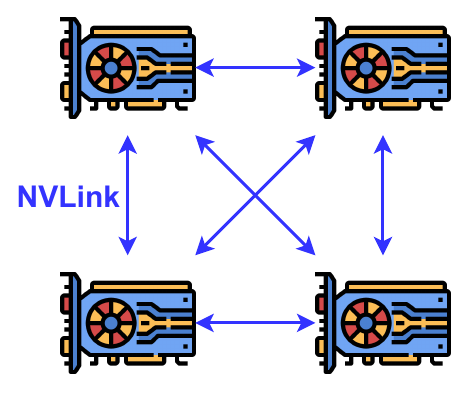}
        \caption{Fully Connected GPUs}
        \label{fig:full-nvlinks}
    \end{subfigure} 
    ~
    \begin{subfigure}{0.2\textwidth}
        \centering
        \includegraphics[width=\linewidth]{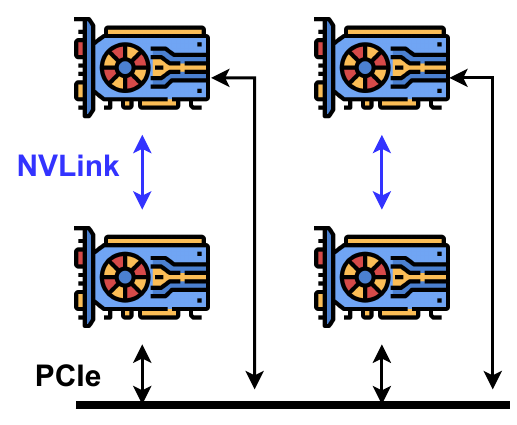}
        \caption{Partially Connected GPUs}
        \label{fig:pair-nvlinks}
    \end{subfigure}
    \caption{Common network topology on GPU nodes}
\end{figure}

3) Hardware Compatibility

\begin{figure}[htp]
    \begin{subfigure}{0.24\textwidth}
        \centering
        \includegraphics[height=1.2in]{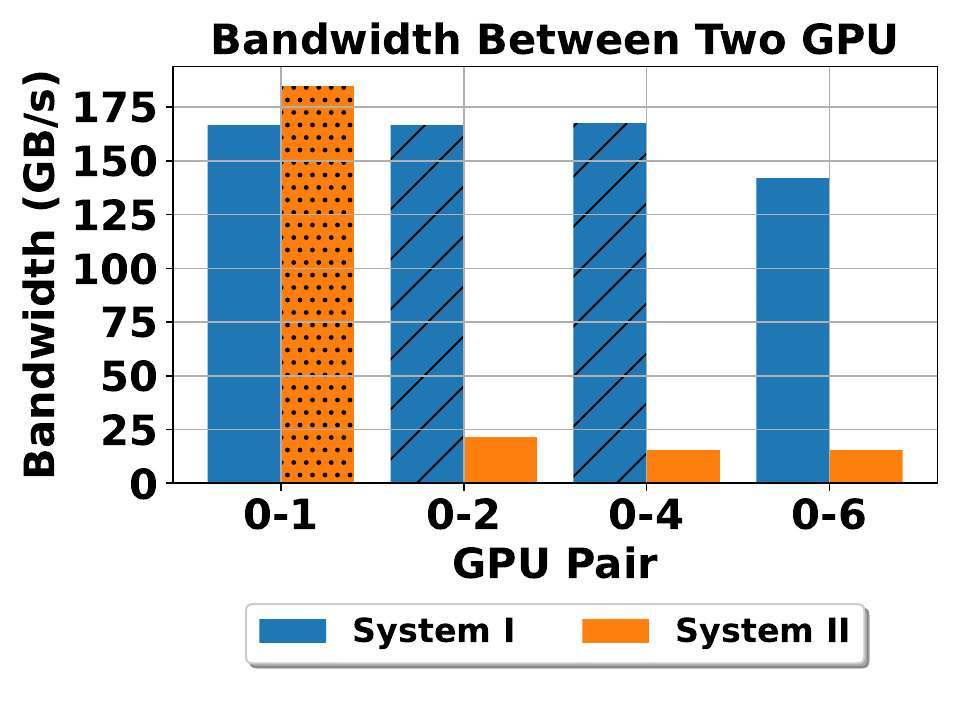}
        \caption{Communication Bandwidth between GPU Pairs}
        \label{fig: bw-gpu-pair}
    \end{subfigure} 
    ~
    \begin{subfigure}{0.24\textwidth}
        \centering
        \includegraphics[height=1.2in]{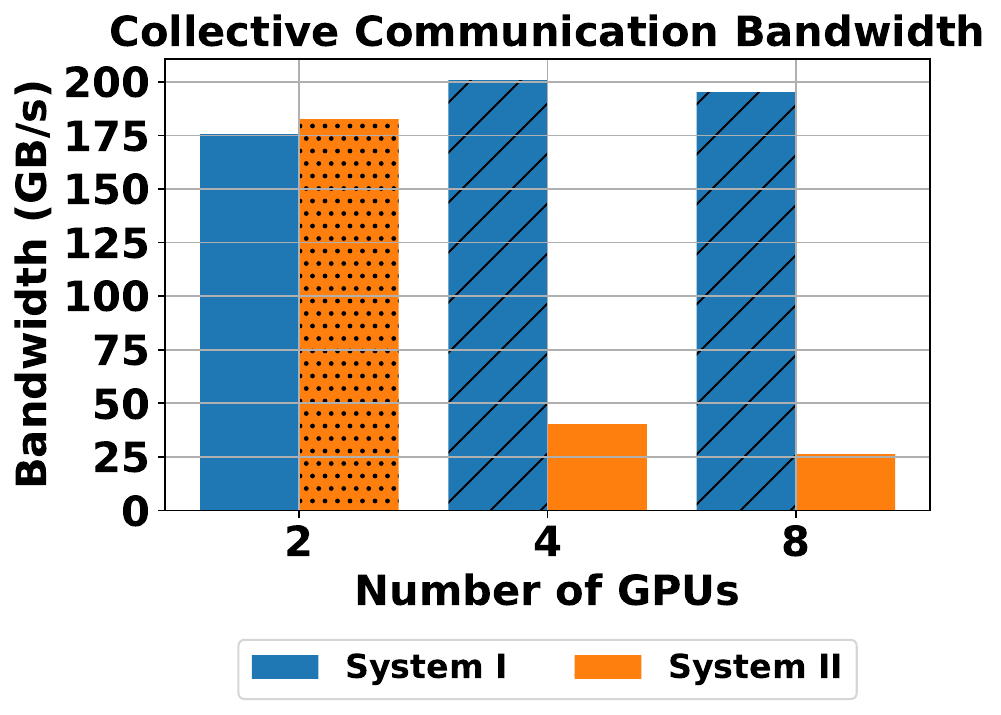}
        \caption{Communication Bandwidth for Collective Communication}
        \label{fig: bw-col-comm}
    \end{subfigure}
    \caption{Communication Bandwidth on System I and II (broadcasting 125 MB data using the NCCL Bandwidth Test tool)}
    \label{fig: bw}
\end{figure}

Experiments were conducted on Systems I and II to further investigate the impact of GPU interconnect on the performance of tensor parallelism. System I and System II were selected for experiments as the former have fully connected NVLink between any pair of GPUs as shown in Figure~\ref{fig:full-nvlinks} while the latter only has NVLink between 4 pairs of adjacent GPUs as shown in Figure~\ref{fig:pair-nvlinks}. The communication bandwidth of System I is consistently high regardless of whether it is measured for a pair of GPUs or a group of GPUs as shown in Figure~\ref{fig: bw}. However, the communication bandwidth drops significantly from 184 GB/s to 15 GB/s when the communication occurs among non-adjacent GPUs as only adjacent GPUs have high-performance NVLink.

The GPU topology of System II is therefore not friendly to 1D tensor parallelism which relies on all-reduce operations across all the GPUs via PCIe. Instead, the 2D and 2.5D only have communication between a pair of GPUs instead of across all GPUs. This allows part of the communication to still utilize the high NVLink bandwidth between adjacent GPUs.

\begin{figure}[]
    \begin{subfigure}{0.24\textwidth}
        \centering
        \includegraphics[height=1.2in]{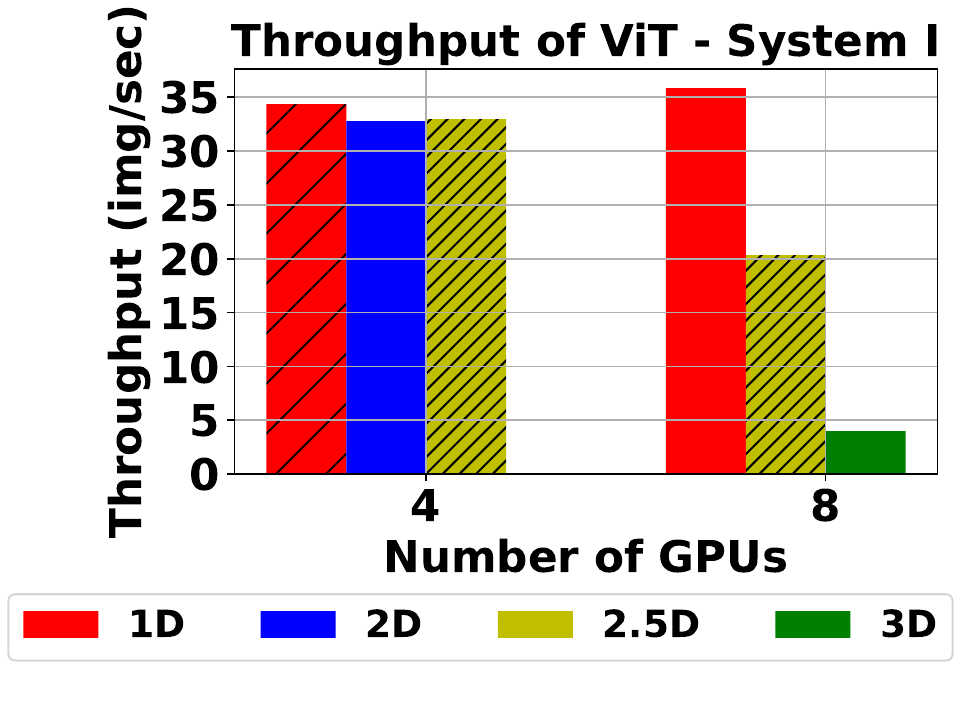}
        \caption{System I}
        \label{fig: throughput-sys-i}
    \end{subfigure} 
    ~
    \begin{subfigure}{0.24\textwidth}
        \centering
        \includegraphics[height=1.2in]{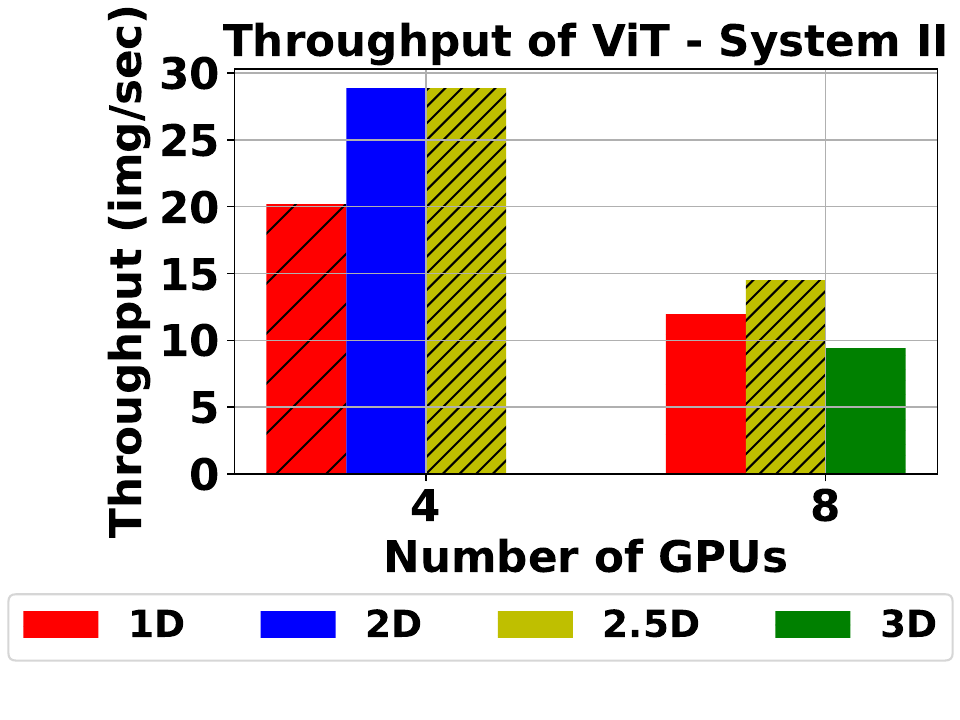}
        \caption{System II}
        \label{fig: throughput-sys-ii}
    \end{subfigure}
    \caption{Throughput of ViT Training on System I and II}
    \label{fig: throughput-vit}
\end{figure}

We trained ViT on the ImageNet-1k dataset with different configurations for 4 GPUs and 8 GPUs on both System I and II. On 4 GPUs, the ViT model has 64 Transformer layers with hidden size of 3072 and 48 attention heads. On 8 GPUs, the hidden size and the number of attention heads are adjusted to 4096 and 64 respectively as there is more memory available. The model is trained with increasing batch size until the out-of-memory problem occurs. As such, we present the best throughput for each tensor parallelism method. In Figure~\ref{fig: throughput-sys-i}, the throughput of 2D, 2.5D, and 3D tensor parallelism cannot compete with 1D tensor parallelism on both 4 GPUs and 8 GPUs. This is expected for two reasons. The first reason is that 1D tensor parallelism can utilize the high communication bandwidth with all GPUs involved in System I. The second reason is that 2D, 2.5D, and 3D tensor parallelisms have more communication volume with a small number of processors and will only surpass 1D tensor parallelism when the number of processors increases.

However, when the experiment is switched to System II in Figure~\ref{fig: throughput-sys-ii}, 1D tensor parallelism will encounter a bottleneck due to the low communication bandwidth in collective communication across 4 and 8 GPUs. Meanwhile, 2D and 2.5D can deliver a throughput that is $40\%$ higher than that of 1D tensor parallelism with 4 GPUs. With 8 GPUs, 2.5D tensor parallelism can still outperform 1D tensor parallelism by $20.6\%$. 3D tensor parallelism still delivers lower performance than 1D tensor parallelism due to the low scaling.

4) Throughput Comparison

To test the performance of tensor parallelism with more GPUs, we trained ViT on System IV. As System IV only has 16 GB GPU memory, therefore, we adjusted the configuration of the ViT model accordingly. The model is set to have 24 layers with the hidden size of 2048 and 32 attention heads for 4 and 8 GPUs. From 16 GPUs onwards, the model is set to have 32 layers with the hidden size of 4096 and 64 attention heads.

The results for 4 to 64 GPUs are shown in Table~\ref{tab: perf-with-gpu}. It can be observed that as the number of GPUs increases, the speedup of advanced tensor parallelism over 1D tensor parallelism increases up to 2.76. This can be attributed to the lower communication volume of advanced tensor parallelism methods when scaling to more processors. Together with memory efficiency and low communication volume, 2D, 2.5D, and 3D tensor parallelism is a better option for large-scale distributed training.

\subsection{Sequence Parallelism}

In this section, we compare Sequence Parallelism with 1D tensor parallelism for memory efficiency and training throughput. As Sequence Parallelism is designed for situations where activations consume more memory than model data, BERT-Base is chosen as our experiment model and trained on the Wikipedia dataset~\cite{wikidump}. We conducted the experiments on System III. It should be noted that 1D tensor parallelism requires the number of attention heads (12) to be divisible by the parallel size, we can only use 4, 6, and 12 GPUs whereas the 6-GPU experiment uses 2 nodes and 3 GPUs from each node. Meanwhile, Sequence Parallelism is not limited by the number of attention heads, thus we conducted experiments on 4, 8, and 12 GPUs.

\begin{figure}[htp]
    \begin{subfigure}{0.24\textwidth}
        \centering
        \includegraphics[height=1.2in]{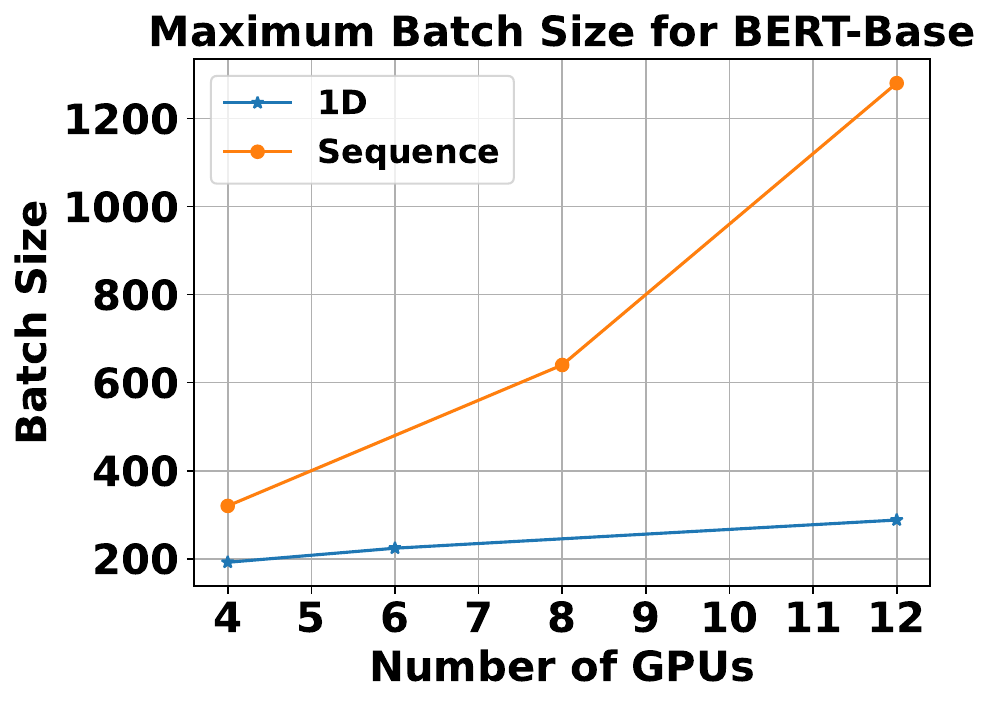}
        \caption{Max Batch Size}
        \label{fig: seq-max-bs}
    \end{subfigure} 
    ~
    \begin{subfigure}{0.24\textwidth}
        \centering
        \includegraphics[height=1.2in]{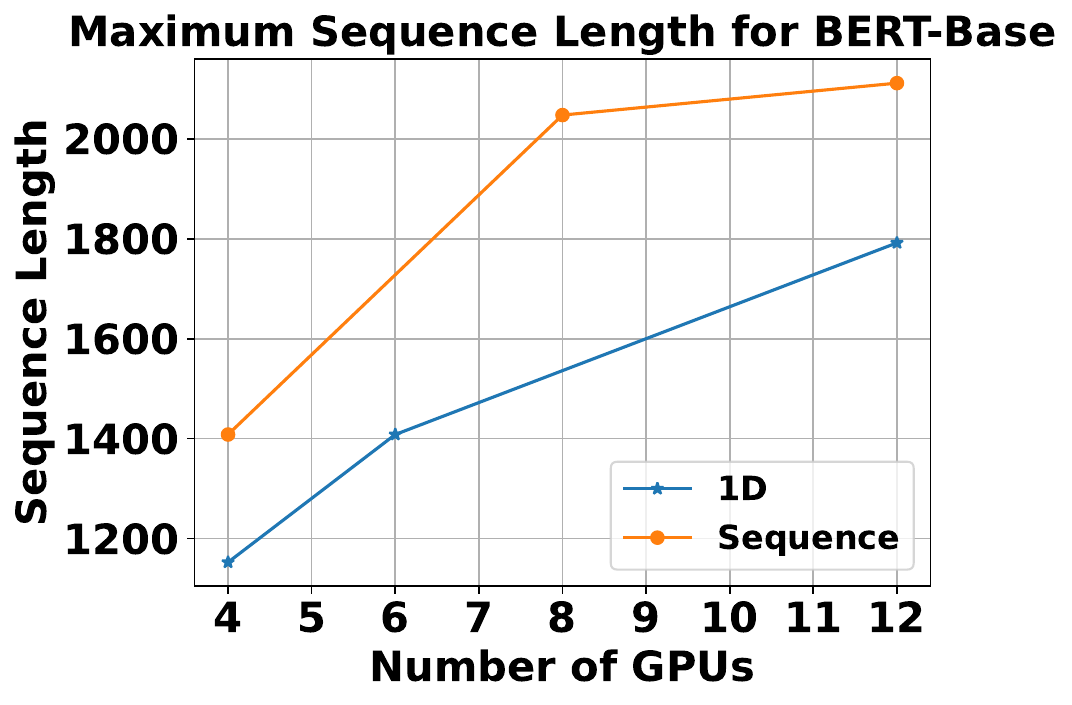}
        \caption{Max Sequence Length}
        \label{fig: seq-max-seq}
    \end{subfigure}
    \caption{Memory Efficiency of Sequence Parallelism over 1D Tensor Parallelism}
    \label{fig: seq-mem}
\end{figure}

1) Memory Efficiency

We increase the batch size and sequence length until the out-of-memory problem occurs for both 1D tensor parallelism and Sequence Parallelism. The sequence length is fixed at 512 for the batch size test while the batch size is fixed at 64 for the sequence length test.

As shown in Figure~\ref{fig: seq-max-bs}, Sequence Parallelism can reach larger batch size than 1D tensor parallelism. This is because that 1D tensor parallelism has a memory bottleneck in the duplicated activations where the activation is split along the sequence dimension in Sequence Parallelism. The maximum batch size of Sequence Parallelism is 4.44 times larger than that of the 1D tensor parallelism with 12 GPUs. 
The same pattern is observed in the maximum sequence length test as shown in Figure~\ref{fig: seq-max-seq}. The maximum sequence length of Sequence Parallelism is 1.18 times larger than that of 1D tensor parallelism. If linear-complexity attention modules~\cite{wang2020linformer, NEURIPS2020_c8512d14} is used instead of the quadratic-complexity self-attention in BERT, Sequence Parallelism can achieve linear scaling of maximum sequence length with the number of GPUs, better supporting document-level text understanding.

\begin{figure}[htp]
    \begin{subfigure}{0.23\textwidth}
        \centering
        \includegraphics[height=1in]{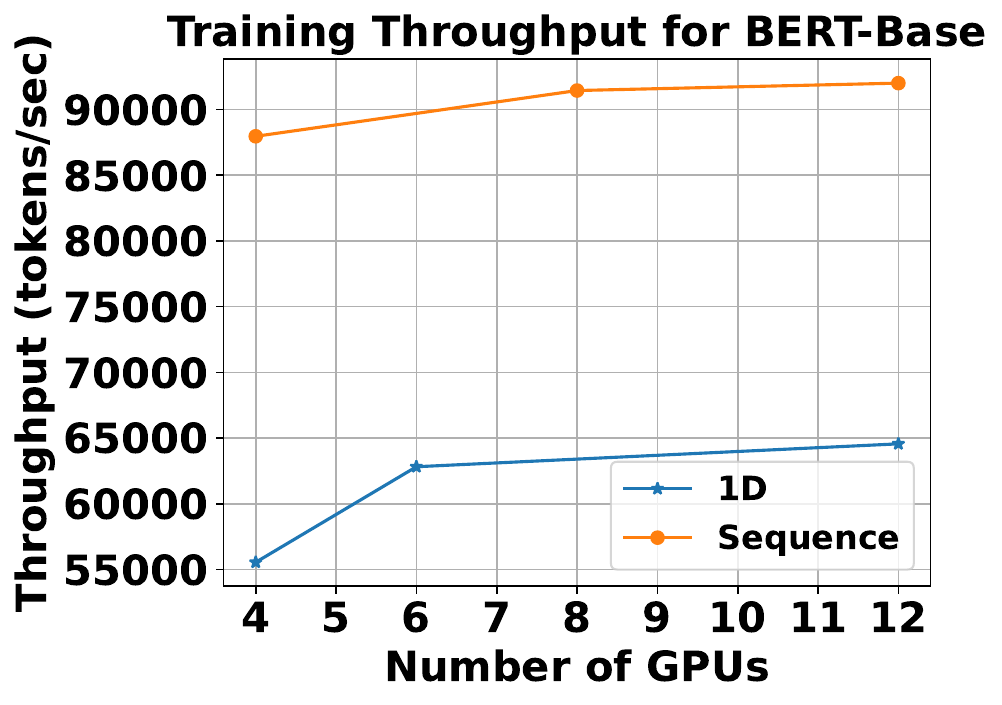}
        \caption{Wihtout Pipeline}
        \label{fig: seq-throughput}
    \end{subfigure} 
    ~
    \begin{subfigure}{0.23\textwidth}
        \centering
        \includegraphics[height=1in]{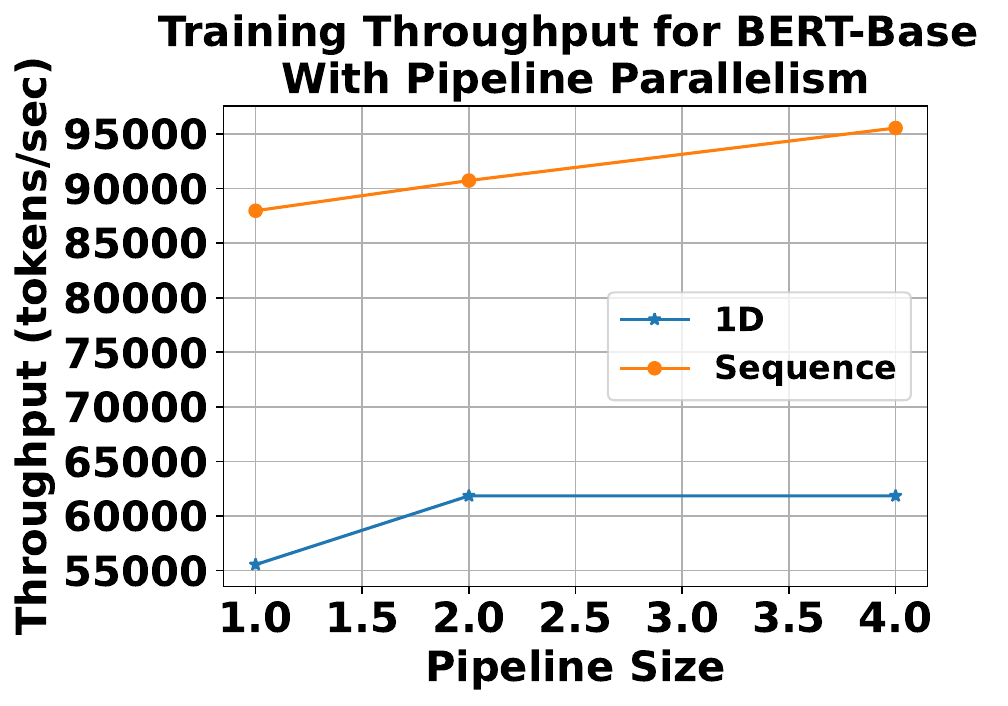}
        \caption{With Pipeline}
        \label{fig: seq-throughput-pipe}
    \end{subfigure}
    \caption{Training Throughput of BERT-Base}
    \label{fig: seq-mem}
\end{figure}

2) Throughput Comparison

To evaluate the training speed of Sequence Parallelism, we trained BERT-Base with the sequence length of 512 and its maximum batch size. As shown in Figure~\ref{fig: seq-throughput}, Sequence Parallelism is up to 1.43 times faster than that of 1D tensor parallelism. 

As sequence parallelism splits the input data and activation, it is naturally compatible with Pipeline Parallelism. While 1D tensor parallelism will split the activation before transferring the tensor to the next stage and gather it back afterward, Sequence Parallelism requires no such communication between pipeline stages. We further scaled the training with Pipeline Parallelism. The parallel size for both Sequence and 1D tensor parallelism is fixed at 4 and we scale the number of pipeline stages from 1 to 4. As shown in Figure~\ref{fig: seq-throughput-pipe}, Sequence Parallelism can train 1.55 times faster than 1D tensor parallelism with 4 pipeline stages.

\subsection{Sharding and Offloading}

\begin{figure}[h]
    \centering
    \includegraphics[width=0.27\textwidth]{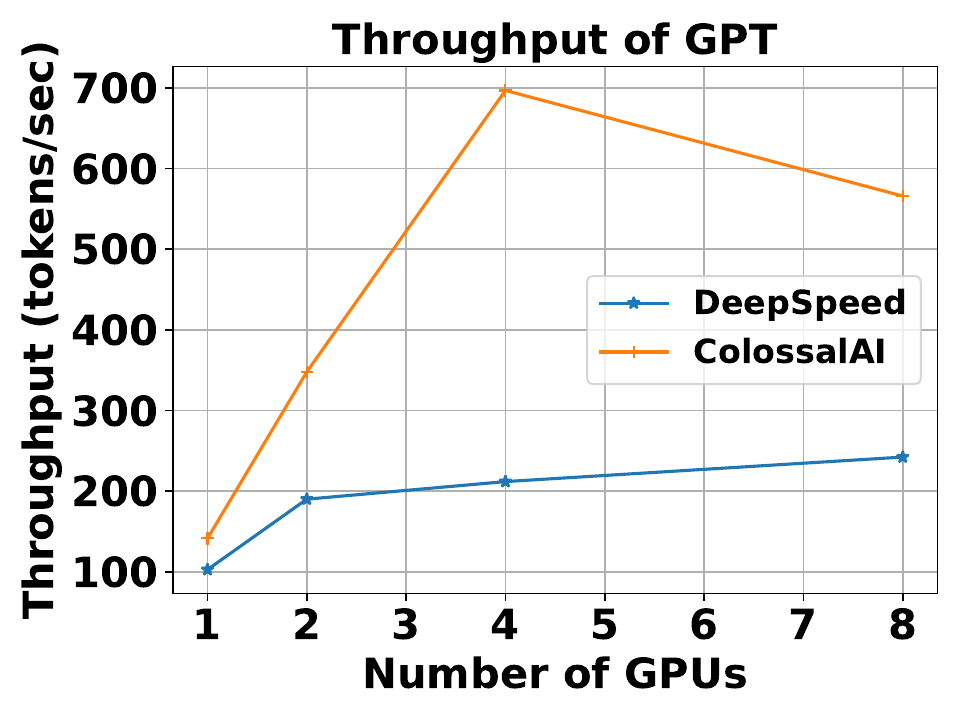}
    \caption{Throughput of GPT Training with Sharding and Offloading with Batch Size 4}
    \label{fig: zero}
\end{figure}

In this section, we evaluated our own sharding and offloading module as discussed in Section ~\ref{sec:sharding&offloading} against DeepSpeed. We used DeepSpeed Stage 3 as the baseline, which partitions model parameters, gradients, and optimizer states in data parallel training. To demonstrate the capability of dynamic tensor placement in ColossalAI, we trained GPT-2 model with 10 billion parameters on the Wikipedia dataset on System II. We set the batch size to 4 and scaled the data parallel training from 1 GPU to 8 GPU. As the batch size is small, the GPU memory is not completely used up. However, DeepSpeed's static policy will still offload all model data to the CPU memory, leading to low memory efficiency and high communication overhead. Instead, Colossal-AI will dynamically determine whether a tensor should be placed on GPU or CPU depending on the memory availability. In this case, since Colossal-AI detects that there is still free memory on the GPU, it will only offload a small portion of the model data, leading to better utilization of the hardware resources and better training throughput as shown in Figure~\ref{fig: zero}. We have also performed training on the OPT model~\cite{zhang2022opt} of 13 billion parameters with the batch size per GPU equal to 32. With a larger batch size, both systems utilized almost all GPU memory and Colossal-AI can still achieve 1.33 times speed up over DeepSpeed on 8 GPUs.

\section{Future Work}

The Colossal-AI system is open-sourced and maintained on GitHub. One future work is to design a hardware-aware and efficient algorithm to automatically search for the optimal parallelization strategy as mentioned in Section ~\ref{sec: automatic-parallelization}. As an open-source project, we would actively integrate with model zoos such as Hugging Face Transformers~\cite{wolf2020huggingfaces}. We expect ColossalAI to be capable of parallelizing models in state-of-the-art model zoos so that distributed training can be more accessible to the Deep Learning community.

\section{Conclusion}

In this work, we designed and implemented Colossal-AI which integrated a vast number of advanced acceleration techniques into one unified system for large-scale distributed training. Colossal-AI comes with a flexible system design that supports an easy combination of different parallelism methods. In addition, its acceleration techniques provide robust performance under different hardware conditions and deliver superior performance compared to the baseline systems. In our experiments, we have demonstrated Colossal-AI can achieve up to 2.76x speedup over the baseline systems.

\bibliographystyle{ACM-Reference-Format}
\bibliography{sample-base}


\begin{thebibliography}{42}


\ifx \showCODEN    \undefined \def \showCODEN     #1{\unskip}     \fi
\ifx \showDOI      \undefined \def \showDOI       #1{#1}\fi
\ifx \showISBNx    \undefined \def \showISBNx     #1{\unskip}     \fi
\ifx \showISBNxiii \undefined \def \showISBNxiii  #1{\unskip}     \fi
\ifx \showISSN     \undefined \def \showISSN      #1{\unskip}     \fi
\ifx \showLCCN     \undefined \def \showLCCN      #1{\unskip}     \fi
\ifx \shownote     \undefined \def \shownote      #1{#1}          \fi
\ifx \showarticletitle \undefined \def \showarticletitle #1{#1}   \fi
\ifx \showURL      \undefined \def \showURL       {\relax}        \fi
\providecommand\bibfield[2]{#2}
\providecommand\bibinfo[2]{#2}
\providecommand\natexlab[1]{#1}
\providecommand\showeprint[2][]{arXiv:#2}

\bibitem[Agarwal et~al\mbox{.}(1995)]%
        {5389455}
\bibfield{author}{\bibinfo{person}{R.~C. Agarwal}, \bibinfo{person}{S.~M.
  Balle}, \bibinfo{person}{F.~G. Gustavson}, \bibinfo{person}{M. Joshi}, {and}
  \bibinfo{person}{P. Palkar}.} \bibinfo{year}{1995}\natexlab{}.
\newblock \showarticletitle{A three-dimensional approach to parallel matrix
  multiplication}.
\newblock \bibinfo{journal}{\emph{IBM Journal of Research and Development}}
  \bibinfo{volume}{39}, \bibinfo{number}{5} (\bibinfo{year}{1995}),
  \bibinfo{pages}{575--582}.
\newblock
\urldef\tempurl%
\url{https://doi.org/10.1147/rd.395.0575}
\showDOI{\tempurl}


\bibitem[Baines et~al\mbox{.}(2021)]%
        {FairScale2021}
\bibfield{author}{\bibinfo{person}{Mandeep Baines}, \bibinfo{person}{Shruti
  Bhosale}, \bibinfo{person}{Vittorio Caggiano}, \bibinfo{person}{Naman Goyal},
  \bibinfo{person}{Siddharth Goyal}, \bibinfo{person}{Myle Ott},
  \bibinfo{person}{Benjamin Lefaudeux}, \bibinfo{person}{Vitaliy Liptchinsky},
  \bibinfo{person}{Mike Rabbat}, \bibinfo{person}{Sam Sheiffer},
  \bibinfo{person}{Anjali Sridhar}, {and} \bibinfo{person}{Min Xu}.}
  \bibinfo{year}{2021}\natexlab{}.
\newblock \bibinfo{title}{FairScale: A general purpose modular PyTorch library
  for high performance and large scale training}.
\newblock
  \bibinfo{howpublished}{\url{https://github.com/facebookresearch/fairscale}}.
\newblock


\bibitem[Berntsen(1989)]%
        {berntsen1989communication}
\bibfield{author}{\bibinfo{person}{Jarle Berntsen}.}
  \bibinfo{year}{1989}\natexlab{}.
\newblock \showarticletitle{Communication efficient matrix multiplication on
  hypercubes}.
\newblock \bibinfo{journal}{\emph{Parallel computing}} \bibinfo{volume}{12},
  \bibinfo{number}{3} (\bibinfo{year}{1989}), \bibinfo{pages}{335--342}.
\newblock


\bibitem[Bian et~al\mbox{.}(2021)]%
        {bian2021_3d}
\bibfield{author}{\bibinfo{person}{Zhengda Bian}, \bibinfo{person}{Qifan Xu},
  \bibinfo{person}{Boxiang Wang}, {and} \bibinfo{person}{Yang You}.}
  \bibinfo{year}{2021}\natexlab{}.
\newblock \showarticletitle{Maximizing Parallelism in Distributed Training for
  Huge Neural Networks}.
\newblock \bibinfo{journal}{\emph{arXiv preprint arXiv:2105.14450}}
  (\bibinfo{year}{2021}).
\newblock


\bibitem[Brown et~al\mbox{.}(2020)]%
        {brown2020language}
\bibfield{author}{\bibinfo{person}{Tom~B Brown}, \bibinfo{person}{Benjamin
  Mann}, \bibinfo{person}{Nick Ryder}, \bibinfo{person}{Melanie Subbiah},
  \bibinfo{person}{Jared Kaplan}, \bibinfo{person}{Prafulla Dhariwal},
  \bibinfo{person}{Arvind Neelakantan}, \bibinfo{person}{Pranav Shyam},
  \bibinfo{person}{Girish Sastry}, \bibinfo{person}{Amanda Askell},
  {et~al\mbox{.}}} \bibinfo{year}{2020}\natexlab{}.
\newblock \showarticletitle{Language models are few-shot learners}.
\newblock \bibinfo{journal}{\emph{arXiv preprint arXiv:2005.14165}}
  (\bibinfo{year}{2020}).
\newblock


\bibitem[Cannon(1969)]%
        {cannon1969cellular}
\bibfield{author}{\bibinfo{person}{Lynn~Elliot Cannon}.}
  \bibinfo{year}{1969}\natexlab{}.
\newblock \bibinfo{booktitle}{\emph{A cellular computer to implement the Kalman
  filter algorithm}}.
\newblock \bibinfo{publisher}{Montana State University}.
\newblock


\bibitem[Chen et~al\mbox{.}(2016)]%
        {act_ckpt}
\bibfield{author}{\bibinfo{person}{Tianqi Chen}, \bibinfo{person}{Bing Xu},
  \bibinfo{person}{Chiyuan Zhang}, {and} \bibinfo{person}{Carlos Guestrin}.}
  \bibinfo{year}{2016}\natexlab{}.
\newblock \bibinfo{title}{Training Deep Nets with Sublinear Memory Cost}.
\newblock
\newblock
\urldef\tempurl%
\url{https://doi.org/10.48550/ARXIV.1604.06174}
\showDOI{\tempurl}


\bibitem[Devlin et~al\mbox{.}(2018)]%
        {devlin2018bert}
\bibfield{author}{\bibinfo{person}{Jacob Devlin}, \bibinfo{person}{Ming-Wei
  Chang}, \bibinfo{person}{Kenton Lee}, {and} \bibinfo{person}{Kristina
  Toutanova}.} \bibinfo{year}{2018}\natexlab{}.
\newblock \showarticletitle{Bert: Pre-training of deep bidirectional
  transformers for language understanding}.
\newblock \bibinfo{journal}{\emph{arXiv preprint arXiv:1810.04805}}
  (\bibinfo{year}{2018}).
\newblock


\bibitem[Dosovitskiy et~al\mbox{.}(2020)]%
        {dosovitskiy2020image}
\bibfield{author}{\bibinfo{person}{Alexey Dosovitskiy}, \bibinfo{person}{Lucas
  Beyer}, \bibinfo{person}{Alexander Kolesnikov}, \bibinfo{person}{Dirk
  Weissenborn}, \bibinfo{person}{Xiaohua Zhai}, \bibinfo{person}{Thomas
  Unterthiner}, \bibinfo{person}{Mostafa Dehghani}, \bibinfo{person}{Matthias
  Minderer}, \bibinfo{person}{Georg Heigold}, \bibinfo{person}{Sylvain Gelly},
  {et~al\mbox{.}}} \bibinfo{year}{2020}\natexlab{}.
\newblock \showarticletitle{An image is worth 16x16 words: Transformers for
  image recognition at scale}.
\newblock \bibinfo{journal}{\emph{arXiv preprint arXiv:2010.11929}}
  (\bibinfo{year}{2020}).
\newblock


\bibitem[Du et~al\mbox{.}(2021)]%
        {du2021all}
\bibfield{author}{\bibinfo{person}{Zhengxiao Du}, \bibinfo{person}{Yujie Qian},
  \bibinfo{person}{Xiao Liu}, \bibinfo{person}{Ming Ding},
  \bibinfo{person}{Jiezhong Qiu}, \bibinfo{person}{Zhilin Yang}, {and}
  \bibinfo{person}{Jie Tang}.} \bibinfo{year}{2021}\natexlab{}.
\newblock \showarticletitle{All NLP Tasks Are Generation Tasks: A General
  Pretraining Framework}.
\newblock \bibinfo{journal}{\emph{arXiv preprint arXiv:2103.10360}}
  (\bibinfo{year}{2021}).
\newblock


\bibitem[Duchi et~al\mbox{.}(2011)]%
        {duchi2011adaptive}
\bibfield{author}{\bibinfo{person}{John Duchi}, \bibinfo{person}{Elad Hazan},
  {and} \bibinfo{person}{Yoram Singer}.} \bibinfo{year}{2011}\natexlab{}.
\newblock \showarticletitle{Adaptive subgradient methods for online learning
  and stochastic optimization.}
\newblock \bibinfo{journal}{\emph{Journal of machine learning research}}
  \bibinfo{volume}{12}, \bibinfo{number}{7} (\bibinfo{year}{2011}).
\newblock


\bibitem[Fang et~al\mbox{.}(2021)]%
        {https://doi.org/10.48550/arxiv.2108.05818}
\bibfield{author}{\bibinfo{person}{Jiarui Fang}, \bibinfo{person}{Yang Yu},
  \bibinfo{person}{Zilin Zhu}, \bibinfo{person}{Shenggui Li},
  \bibinfo{person}{Yang You}, {and} \bibinfo{person}{Jie Zhou}.}
  \bibinfo{year}{2021}\natexlab{}.
\newblock \bibinfo{title}{PatrickStar: Parallel Training of Pre-trained Models
  via Chunk-based Memory Management}.
\newblock
\newblock
\urldef\tempurl%
\url{https://doi.org/10.48550/ARXIV.2108.05818}
\showDOI{\tempurl}


\bibitem[Foundation({[n.\,d.]})]%
        {wikidump}
\bibfield{author}{\bibinfo{person}{Wikimedia Foundation}.}
  \bibinfo{year}{[n.\,d.]}\natexlab{}.
\newblock \bibinfo{booktitle}{\emph{Wikimedia Downloads}}.
\newblock
\urldef\tempurl%
\url{https://dumps.wikimedia.org}
\showURL{%
\tempurl}


\bibitem[He et~al\mbox{.}(2016)]%
        {resnet}
\bibfield{author}{\bibinfo{person}{Kaiming He}, \bibinfo{person}{Xiangyu
  Zhang}, \bibinfo{person}{Shaoqing Ren}, {and} \bibinfo{person}{Jian Sun}.}
  \bibinfo{year}{2016}\natexlab{}.
\newblock \showarticletitle{Deep Residual Learning for Image Recognition}. In
  \bibinfo{booktitle}{\emph{2016 IEEE Conference on Computer Vision and Pattern
  Recognition (CVPR)}}. \bibinfo{pages}{770--778}.
\newblock
\urldef\tempurl%
\url{https://doi.org/10.1109/CVPR.2016.90}
\showDOI{\tempurl}


\bibitem[Huang et~al\mbox{.}(2019a)]%
        {NEURIPS2019_093f65e0}
\bibfield{author}{\bibinfo{person}{Yanping Huang}, \bibinfo{person}{Youlong
  Cheng}, \bibinfo{person}{Ankur Bapna}, \bibinfo{person}{Orhan Firat},
  \bibinfo{person}{Dehao Chen}, \bibinfo{person}{Mia Chen},
  \bibinfo{person}{HyoukJoong Lee}, \bibinfo{person}{Jiquan Ngiam},
  \bibinfo{person}{Quoc~V Le}, \bibinfo{person}{Yonghui Wu}, {and}
  \bibinfo{person}{zhifeng Chen}.} \bibinfo{year}{2019}\natexlab{a}.
\newblock \showarticletitle{GPipe: Efficient Training of Giant Neural Networks
  using Pipeline Parallelism}. In \bibinfo{booktitle}{\emph{Advances in Neural
  Information Processing Systems}},
  \bibfield{editor}{\bibinfo{person}{H.~Wallach},
  \bibinfo{person}{H.~Larochelle}, \bibinfo{person}{A.~Beygelzimer},
  \bibinfo{person}{F.~d\textquotesingle Alch\'{e}-Buc},
  \bibinfo{person}{E.~Fox}, {and} \bibinfo{person}{R.~Garnett}} (Eds.),
  Vol.~\bibinfo{volume}{32}. \bibinfo{publisher}{Curran Associates, Inc.}
\newblock
\urldef\tempurl%
\url{https://proceedings.neurips.cc/paper/2019/file/093f65e080a295f8076b1c5722a46aa2-Paper.pdf}
\showURL{%
\tempurl}


\bibitem[Huang et~al\mbox{.}(2019b)]%
        {gpipe}
\bibfield{author}{\bibinfo{person}{Yanping Huang}, \bibinfo{person}{Youlong
  Cheng}, \bibinfo{person}{Ankur Bapna}, \bibinfo{person}{Orhan Firat},
  \bibinfo{person}{Mia~Xu Chen}, \bibinfo{person}{Dehao Chen},
  \bibinfo{person}{HyoukJoong Lee}, \bibinfo{person}{Jiquan Ngiam},
  \bibinfo{person}{Quoc~V. Le}, \bibinfo{person}{Yonghui Wu}, {and}
  \bibinfo{person}{Zhifeng Chen}.} \bibinfo{year}{2019}\natexlab{b}.
\newblock \bibinfo{booktitle}{\emph{GPipe: Efficient Training of Giant Neural
  Networks Using Pipeline Parallelism}}.
\newblock \bibinfo{publisher}{Curran Associates Inc.}, \bibinfo{address}{Red
  Hook, NY, USA}.
\newblock


\bibitem[Kingma and Ba(2014a)]%
        {adamoptim}
\bibfield{author}{\bibinfo{person}{Diederik Kingma} {and}
  \bibinfo{person}{Jimmy Ba}.} \bibinfo{year}{2014}\natexlab{a}.
\newblock \showarticletitle{Adam: A Method for Stochastic Optimization}.
\newblock \bibinfo{journal}{\emph{International Conference on Learning
  Representations}} (\bibinfo{date}{12} \bibinfo{year}{2014}).
\newblock


\bibitem[Kingma and Ba(2014b)]%
        {kingma2014adam}
\bibfield{author}{\bibinfo{person}{Diederik~P Kingma} {and}
  \bibinfo{person}{Jimmy Ba}.} \bibinfo{year}{2014}\natexlab{b}.
\newblock \showarticletitle{Adam: A method for stochastic optimization}.
\newblock \bibinfo{journal}{\emph{arXiv preprint arXiv:1412.6980}}
  (\bibinfo{year}{2014}).
\newblock


\bibitem[Lepikhin et~al\mbox{.}(2021)]%
        {lepikhin2021gshard}
\bibfield{author}{\bibinfo{person}{Dmitry Lepikhin},
  \bibinfo{person}{HyoukJoong Lee}, \bibinfo{person}{Yuanzhong Xu},
  \bibinfo{person}{Dehao Chen}, \bibinfo{person}{Orhan Firat},
  \bibinfo{person}{Yanping Huang}, \bibinfo{person}{Maxim Krikun},
  \bibinfo{person}{Noam Shazeer}, {and} \bibinfo{person}{Zhifeng Chen}.}
  \bibinfo{year}{2021}\natexlab{}.
\newblock \showarticletitle{{\{}GS{\}}hard: Scaling Giant Models with
  Conditional Computation and Automatic Sharding}. In
  \bibinfo{booktitle}{\emph{International Conference on Learning
  Representations}}.
\newblock
\urldef\tempurl%
\url{https://openreview.net/forum?id=qrwe7XHTmYb}
\showURL{%
\tempurl}


\bibitem[Li and Hoefler(2021)]%
        {chimera}
\bibfield{author}{\bibinfo{person}{Shigang Li} {and} \bibinfo{person}{Torsten
  Hoefler}.} \bibinfo{year}{2021}\natexlab{}.
\newblock \showarticletitle{Chimera: Efficiently Training Large-Scale Neural
  Networks with Bidirectional Pipelines}. In
  \bibinfo{booktitle}{\emph{Proceedings of the International Conference for
  High Performance Computing, Networking, Storage and Analysis}} (St. Louis,
  Missouri) \emph{(\bibinfo{series}{SC '21})}. \bibinfo{publisher}{Association
  for Computing Machinery}, \bibinfo{address}{New York, NY, USA}, Article
  \bibinfo{articleno}{27}, \bibinfo{numpages}{14}~pages.
\newblock
\showISBNx{9781450384421}
\urldef\tempurl%
\url{https://doi.org/10.1145/3458817.3476145}
\showDOI{\tempurl}


\bibitem[Li et~al\mbox{.}(2021)]%
        {seq_parallel}
\bibfield{author}{\bibinfo{person}{Shenggui Li}, \bibinfo{person}{Fuzhao Xue},
  \bibinfo{person}{Yongbin Li}, {and} \bibinfo{person}{Yang You}.}
  \bibinfo{year}{2021}\natexlab{}.
\newblock \bibinfo{title}{Sequence Parallelism: Long Sequence Training from
  System Perspective}.
\newblock
\newblock
\urldef\tempurl%
\url{https://doi.org/10.48550/ARXIV.2105.13120}
\showDOI{\tempurl}


\bibitem[Li et~al\mbox{.}(2020)]%
        {10.14778/3415478.3415530}
\bibfield{author}{\bibinfo{person}{Shen Li}, \bibinfo{person}{Yanli Zhao},
  \bibinfo{person}{Rohan Varma}, \bibinfo{person}{Omkar Salpekar},
  \bibinfo{person}{Pieter Noordhuis}, \bibinfo{person}{Teng Li},
  \bibinfo{person}{Adam Paszke}, \bibinfo{person}{Jeff Smith},
  \bibinfo{person}{Brian Vaughan}, \bibinfo{person}{Pritam Damania}, {and}
  \bibinfo{person}{Soumith Chintala}.} \bibinfo{year}{2020}\natexlab{}.
\newblock \showarticletitle{PyTorch Distributed: Experiences on Accelerating
  Data Parallel Training}.
\newblock \bibinfo{journal}{\emph{Proc. VLDB Endow.}} \bibinfo{volume}{13},
  \bibinfo{number}{12} (\bibinfo{date}{aug} \bibinfo{year}{2020}),
  \bibinfo{pages}{3005–3018}.
\newblock
\showISSN{2150-8097}
\urldef\tempurl%
\url{https://doi.org/10.14778/3415478.3415530}
\showDOI{\tempurl}


\bibitem[Lu et~al\mbox{.}(2017)]%
        {flexflow}
\bibfield{author}{\bibinfo{person}{Wenyan Lu}, \bibinfo{person}{Guihai Yan},
  \bibinfo{person}{Jiajun Li}, \bibinfo{person}{Shijun Gong},
  \bibinfo{person}{Yinhe Han}, {and} \bibinfo{person}{Xiaowei Li}.}
  \bibinfo{year}{2017}\natexlab{}.
\newblock \showarticletitle{FlexFlow: A Flexible Dataflow Accelerator
  Architecture for Convolutional Neural Networks}. In
  \bibinfo{booktitle}{\emph{2017 IEEE International Symposium on High
  Performance Computer Architecture (HPCA)}}. \bibinfo{pages}{553--564}.
\newblock
\urldef\tempurl%
\url{https://doi.org/10.1109/HPCA.2017.29}
\showDOI{\tempurl}


\bibitem[Narayanan et~al\mbox{.}(2019a)]%
        {10.1145/3341301.3359646}
\bibfield{author}{\bibinfo{person}{Deepak Narayanan}, \bibinfo{person}{Aaron
  Harlap}, \bibinfo{person}{Amar Phanishayee}, \bibinfo{person}{Vivek
  Seshadri}, \bibinfo{person}{Nikhil~R. Devanur}, \bibinfo{person}{Gregory~R.
  Ganger}, \bibinfo{person}{Phillip~B. Gibbons}, {and} \bibinfo{person}{Matei
  Zaharia}.} \bibinfo{year}{2019}\natexlab{a}.
\newblock \showarticletitle{PipeDream: Generalized Pipeline Parallelism for DNN
  Training}. In \bibinfo{booktitle}{\emph{Proceedings of the 27th ACM Symposium
  on Operating Systems Principles}} (Huntsville, Ontario, Canada)
  \emph{(\bibinfo{series}{SOSP '19})}. \bibinfo{publisher}{Association for
  Computing Machinery}, \bibinfo{address}{New York, NY, USA},
  \bibinfo{pages}{1–15}.
\newblock
\showISBNx{9781450368735}
\urldef\tempurl%
\url{https://doi.org/10.1145/3341301.3359646}
\showDOI{\tempurl}


\bibitem[Narayanan et~al\mbox{.}(2019b)]%
        {pipedream}
\bibfield{author}{\bibinfo{person}{Deepak Narayanan}, \bibinfo{person}{Aaron
  Harlap}, \bibinfo{person}{Amar Phanishayee}, \bibinfo{person}{Vivek
  Seshadri}, \bibinfo{person}{Nikhil~R. Devanur}, \bibinfo{person}{Gregory~R.
  Ganger}, \bibinfo{person}{Phillip~B. Gibbons}, {and} \bibinfo{person}{Matei
  Zaharia}.} \bibinfo{year}{2019}\natexlab{b}.
\newblock \showarticletitle{PipeDream: Generalized Pipeline Parallelism for DNN
  Training}. In \bibinfo{booktitle}{\emph{Proceedings of the 27th ACM Symposium
  on Operating Systems Principles}} (Huntsville, Ontario, Canada)
  \emph{(\bibinfo{series}{SOSP '19})}. \bibinfo{publisher}{Association for
  Computing Machinery}, \bibinfo{address}{New York, NY, USA},
  \bibinfo{pages}{1–15}.
\newblock
\showISBNx{9781450368735}
\urldef\tempurl%
\url{https://doi.org/10.1145/3341301.3359646}
\showDOI{\tempurl}


\bibitem[Narayanan et~al\mbox{.}(2021)]%
        {megatron}
\bibfield{author}{\bibinfo{person}{Deepak Narayanan}, \bibinfo{person}{Mohammad
  Shoeybi}, \bibinfo{person}{Jared Casper}, \bibinfo{person}{Patrick
  LeGresley}, \bibinfo{person}{Mostofa Patwary}, \bibinfo{person}{Vijay
  Korthikanti}, \bibinfo{person}{Dmitri Vainbrand}, \bibinfo{person}{Prethvi
  Kashinkunti}, \bibinfo{person}{Julie Bernauer}, \bibinfo{person}{Bryan
  Catanzaro}, \bibinfo{person}{Amar Phanishayee}, {and} \bibinfo{person}{Matei
  Zaharia}.} \bibinfo{year}{2021}\natexlab{}.
\newblock \showarticletitle{Efficient Large-Scale Language Model Training on
  GPU Clusters Using Megatron-LM}. In \bibinfo{booktitle}{\emph{Proceedings of
  the International Conference for High Performance Computing, Networking,
  Storage and Analysis}} (St. Louis, Missouri) \emph{(\bibinfo{series}{SC
  '21})}. \bibinfo{publisher}{Association for Computing Machinery},
  \bibinfo{address}{New York, NY, USA}, Article \bibinfo{articleno}{58},
  \bibinfo{numpages}{15}~pages.
\newblock
\showISBNx{9781450384421}
\urldef\tempurl%
\url{https://doi.org/10.1145/3458817.3476209}
\showDOI{\tempurl}


\bibitem[Paperno et~al\mbox{.}(2016)]%
        {lambada}
\bibfield{author}{\bibinfo{person}{Denis Paperno}, \bibinfo{person}{Germán
  Kruszewski}, \bibinfo{person}{Angeliki Lazaridou}, \bibinfo{person}{Quan
  Pham}, \bibinfo{person}{Raffaella Bernardi}, \bibinfo{person}{Sandro
  Pezzelle}, \bibinfo{person}{Marco Baroni}, \bibinfo{person}{Gemma Boleda},
  {and} \bibinfo{person}{Raquel Fernández}.} \bibinfo{year}{2016}\natexlab{}.
\newblock \showarticletitle{The LAMBADA dataset: Word prediction requiring a
  broad discourse context}. \bibinfo{pages}{1525--1534}.
\newblock
\urldef\tempurl%
\url{https://doi.org/10.18653/v1/P16-1144}
\showDOI{\tempurl}


\bibitem[Radford et~al\mbox{.}(2019)]%
        {Radford2019LanguageMA}
\bibfield{author}{\bibinfo{person}{Alec Radford}, \bibinfo{person}{Jeff Wu},
  \bibinfo{person}{Rewon Child}, \bibinfo{person}{David Luan},
  \bibinfo{person}{Dario Amodei}, {and} \bibinfo{person}{Ilya Sutskever}.}
  \bibinfo{year}{2019}\natexlab{}.
\newblock \showarticletitle{Language Models are Unsupervised Multitask
  Learners}.
\newblock


\bibitem[Rasley et~al\mbox{.}(2020)]%
        {rasley2020deepspeed}
\bibfield{author}{\bibinfo{person}{Jeff Rasley}, \bibinfo{person}{Samyam
  Rajbhandari}, \bibinfo{person}{Olatunji Ruwase}, {and}
  \bibinfo{person}{Yuxiong He}.} \bibinfo{year}{2020}\natexlab{}.
\newblock \showarticletitle{Deepspeed: System optimizations enable training
  deep learning models with over 100 billion parameters}. In
  \bibinfo{booktitle}{\emph{Proceedings of the 26th ACM SIGKDD International
  Conference on Knowledge Discovery \& Data Mining}}.
  \bibinfo{pages}{3505--3506}.
\newblock


\bibitem[Ren et~al\mbox{.}(2021)]%
        {ren2021zerooffload}
\bibfield{author}{\bibinfo{person}{Jie Ren}, \bibinfo{person}{Samyam
  Rajbhandari}, \bibinfo{person}{Reza~Yazdani Aminabadi},
  \bibinfo{person}{Olatunji Ruwase}, \bibinfo{person}{Shuangyan Yang},
  \bibinfo{person}{Minjia Zhang}, \bibinfo{person}{Dong Li}, {and}
  \bibinfo{person}{Yuxiong He}.} \bibinfo{year}{2021}\natexlab{}.
\newblock \bibinfo{title}{ZeRO-Offload: Democratizing Billion-Scale Model
  Training}.
\newblock
\newblock
\showeprint[arxiv]{2101.06840}~[cs.DC]


\bibitem[Sergeev and Balso(2018)]%
        {horovod}
\bibfield{author}{\bibinfo{person}{Alexander Sergeev} {and}
  \bibinfo{person}{Mike~Del Balso}.} \bibinfo{year}{2018}\natexlab{}.
\newblock \showarticletitle{Horovod: fast and easy distributed deep learning in
  TensorFlow}.
\newblock \bibinfo{journal}{\emph{CoRR}}  \bibinfo{volume}{abs/1802.05799}
  (\bibinfo{year}{2018}).
\newblock
\showeprint[arXiv]{1802.05799}
\urldef\tempurl%
\url{http://arxiv.org/abs/1802.05799}
\showURL{%
\tempurl}


\bibitem[Shoeybi et~al\mbox{.}(2019)]%
        {shoeybi2019megatron}
\bibfield{author}{\bibinfo{person}{Mohammad Shoeybi}, \bibinfo{person}{Mostofa
  Patwary}, \bibinfo{person}{Raul Puri}, \bibinfo{person}{Patrick LeGresley},
  \bibinfo{person}{Jared Casper}, {and} \bibinfo{person}{Bryan Catanzaro}.}
  \bibinfo{year}{2019}\natexlab{}.
\newblock \showarticletitle{Megatron-lm: Training multi-billion parameter
  language models using model parallelism}.
\newblock \bibinfo{journal}{\emph{arXiv preprint arXiv:1909.08053}}
  (\bibinfo{year}{2019}).
\newblock


\bibitem[Solomonik and Demmel(2011)]%
        {Solomonik2011CommunicationOptimalP2}
\bibfield{author}{\bibinfo{person}{Edgar Solomonik} {and}
  \bibinfo{person}{James Demmel}.} \bibinfo{year}{2011}\natexlab{}.
\newblock \showarticletitle{Communication-Optimal Parallel 2.5D Matrix
  Multiplication and LU Factorization Algorithms}. In
  \bibinfo{booktitle}{\emph{Euro-Par}}.
\newblock


\bibitem[van~de Geijn and Watts(1995)]%
        {10.5555/899248}
\bibfield{author}{\bibinfo{person}{Robert~A. van~de Geijn} {and}
  \bibinfo{person}{Jerrell Watts}.} \bibinfo{year}{1995}\natexlab{}.
\newblock \bibinfo{booktitle}{\emph{SUMMA: Scalable Universal Matrix
  Multiplication Algorithm}}.
\newblock \bibinfo{type}{{T}echnical {R}eport}. \bibinfo{address}{USA}.
\newblock


\bibitem[Vaswani et~al\mbox{.}(2017)]%
        {attnisalluneed}
\bibfield{author}{\bibinfo{person}{Ashish Vaswani}, \bibinfo{person}{Noam
  Shazeer}, \bibinfo{person}{Niki Parmar}, \bibinfo{person}{Jakob Uszkoreit},
  \bibinfo{person}{Llion Jones}, \bibinfo{person}{Aidan~N Gomez},
  \bibinfo{person}{\L~ukasz Kaiser}, {and} \bibinfo{person}{Illia Polosukhin}.}
  \bibinfo{year}{2017}\natexlab{}.
\newblock \showarticletitle{Attention is All you Need}. In
  \bibinfo{booktitle}{\emph{Advances in Neural Information Processing
  Systems}}, \bibfield{editor}{\bibinfo{person}{I.~Guyon},
  \bibinfo{person}{U.~V. Luxburg}, \bibinfo{person}{S.~Bengio},
  \bibinfo{person}{H.~Wallach}, \bibinfo{person}{R.~Fergus},
  \bibinfo{person}{S.~Vishwanathan}, {and} \bibinfo{person}{R.~Garnett}}
  (Eds.), Vol.~\bibinfo{volume}{30}. \bibinfo{publisher}{Curran Associates,
  Inc.}
\newblock
\urldef\tempurl%
\url{https://proceedings.neurips.cc/paper/2017/file/3f5ee243547dee91fbd053c1c4a845aa-Paper.pdf}
\showURL{%
\tempurl}


\bibitem[Wang et~al\mbox{.}(2021)]%
        {wang_2p5d}
\bibfield{author}{\bibinfo{person}{Boxiang Wang}, \bibinfo{person}{Qifan Xu},
  \bibinfo{person}{Zhengda Bian}, {and} \bibinfo{person}{Yang You}.}
  \bibinfo{year}{2021}\natexlab{}.
\newblock \showarticletitle{2.5-dimensional distributed model training}.
\newblock \bibinfo{journal}{\emph{arXiv preprint arXiv:2105.14500}}
  (\bibinfo{year}{2021}).
\newblock


\bibitem[Wang et~al\mbox{.}(2020)]%
        {wang2020linformer}
\bibfield{author}{\bibinfo{person}{Sinong Wang}, \bibinfo{person}{Belinda Li},
  \bibinfo{person}{Madian Khabsa}, \bibinfo{person}{Han Fang}, {and}
  \bibinfo{person}{Hao Ma}.} \bibinfo{year}{2020}\natexlab{}.
\newblock \showarticletitle{Linformer: Self-Attention with Linear Complexity}.
\newblock \bibinfo{journal}{\emph{arXiv preprint arXiv:2006.04768}}
  (\bibinfo{year}{2020}).
\newblock


\bibitem[Wolf et~al\mbox{.}(2020)]%
        {wolf2020huggingfaces}
\bibfield{author}{\bibinfo{person}{Thomas Wolf}, \bibinfo{person}{Lysandre
  Debut}, \bibinfo{person}{Victor Sanh}, \bibinfo{person}{Julien Chaumond},
  \bibinfo{person}{Clement Delangue}, \bibinfo{person}{Anthony Moi},
  \bibinfo{person}{Pierric Cistac}, \bibinfo{person}{Tim Rault},
  \bibinfo{person}{Rémi Louf}, \bibinfo{person}{Morgan Funtowicz},
  \bibinfo{person}{Joe Davison}, \bibinfo{person}{Sam Shleifer},
  \bibinfo{person}{Patrick von Platen}, \bibinfo{person}{Clara Ma},
  \bibinfo{person}{Yacine Jernite}, \bibinfo{person}{Julien Plu},
  \bibinfo{person}{Canwen Xu}, \bibinfo{person}{Teven~Le Scao},
  \bibinfo{person}{Sylvain Gugger}, \bibinfo{person}{Mariama Drame},
  \bibinfo{person}{Quentin Lhoest}, {and} \bibinfo{person}{Alexander~M. Rush}.}
  \bibinfo{year}{2020}\natexlab{}.
\newblock \bibinfo{title}{HuggingFace's Transformers: State-of-the-art Natural
  Language Processing}.
\newblock
\newblock
\showeprint[arxiv]{1910.03771}~[cs.CL]


\bibitem[Xu et~al\mbox{.}(2021)]%
        {xu2021_2d}
\bibfield{author}{\bibinfo{person}{Qifan Xu}, \bibinfo{person}{Shenggui Li},
  \bibinfo{person}{Chaoyu Gong}, {and} \bibinfo{person}{Yang You}.}
  \bibinfo{year}{2021}\natexlab{}.
\newblock \showarticletitle{An Efficient 2D Method for Training Super-Large
  Deep Learning Models}.
\newblock \bibinfo{journal}{\emph{arXiv preprint arXiv:2104.05343}}
  (\bibinfo{year}{2021}).
\newblock


\bibitem[Zaheer et~al\mbox{.}(2020)]%
        {NEURIPS2020_c8512d14}
\bibfield{author}{\bibinfo{person}{Manzil Zaheer}, \bibinfo{person}{Guru
  Guruganesh}, \bibinfo{person}{Kumar~Avinava Dubey}, \bibinfo{person}{Joshua
  Ainslie}, \bibinfo{person}{Chris Alberti}, \bibinfo{person}{Santiago
  Ontanon}, \bibinfo{person}{Philip Pham}, \bibinfo{person}{Anirudh Ravula},
  \bibinfo{person}{Qifan Wang}, \bibinfo{person}{Li Yang}, {and}
  \bibinfo{person}{Amr Ahmed}.} \bibinfo{year}{2020}\natexlab{}.
\newblock \showarticletitle{Big Bird: Transformers for Longer Sequences}. In
  \bibinfo{booktitle}{\emph{Advances in Neural Information Processing
  Systems}}, \bibfield{editor}{\bibinfo{person}{H.~Larochelle},
  \bibinfo{person}{M.~Ranzato}, \bibinfo{person}{R.~Hadsell},
  \bibinfo{person}{M.~F. Balcan}, {and} \bibinfo{person}{H.~Lin}} (Eds.),
  Vol.~\bibinfo{volume}{33}. \bibinfo{publisher}{Curran Associates, Inc.},
  \bibinfo{pages}{17283--17297}.
\newblock
\urldef\tempurl%
\url{https://proceedings.neurips.cc/paper/2020/file/c8512d142a2d849725f31a9a7a361ab9-Paper.pdf}
\showURL{%
\tempurl}


\bibitem[Zhang et~al\mbox{.}(2022)]%
        {zhang2022opt}
\bibfield{author}{\bibinfo{person}{Susan Zhang}, \bibinfo{person}{Stephen
  Roller}, \bibinfo{person}{Naman Goyal}, \bibinfo{person}{Mikel Artetxe},
  \bibinfo{person}{Moya Chen}, \bibinfo{person}{Shuohui Chen},
  \bibinfo{person}{Christopher Dewan}, \bibinfo{person}{Mona Diab},
  \bibinfo{person}{Xian Li}, \bibinfo{person}{Xi~Victoria Lin},
  \bibinfo{person}{Todor Mihaylov}, \bibinfo{person}{Myle Ott},
  \bibinfo{person}{Sam Shleifer}, \bibinfo{person}{Kurt Shuster},
  \bibinfo{person}{Daniel Simig}, \bibinfo{person}{Punit~Singh Koura},
  \bibinfo{person}{Anjali Sridhar}, \bibinfo{person}{Tianlu Wang}, {and}
  \bibinfo{person}{Luke Zettlemoyer}.} \bibinfo{year}{2022}\natexlab{}.
\newblock \bibinfo{title}{OPT: Open Pre-trained Transformer Language Models}.
\newblock
\newblock
\showeprint[arxiv]{2205.01068}~[cs.CL]


\bibitem[Zheng et~al\mbox{.}(2022)]%
        {alpa}
\bibfield{author}{\bibinfo{person}{Lianmin Zheng}, \bibinfo{person}{Zhuohan
  Li}, \bibinfo{person}{Hao Zhang}, \bibinfo{person}{Yonghao Zhuang},
  \bibinfo{person}{Zhifeng Chen}, \bibinfo{person}{Yanping Huang},
  \bibinfo{person}{Yida Wang}, \bibinfo{person}{Yuanzhong Xu},
  \bibinfo{person}{Danyang Zhuo}, \bibinfo{person}{Eric~P. Xing},
  \bibinfo{person}{Joseph~E. Gonzalez}, {and} \bibinfo{person}{Ion Stoica}.}
  \bibinfo{year}{2022}\natexlab{}.
\newblock \showarticletitle{Alpa: Automating Inter- and {Intra-Operator}
  Parallelism for Distributed Deep Learning}. In \bibinfo{booktitle}{\emph{16th
  USENIX Symposium on Operating Systems Design and Implementation (OSDI 22)}}.
  \bibinfo{publisher}{USENIX Association}, \bibinfo{address}{Carlsbad, CA},
  \bibinfo{pages}{559--578}.
\newblock
\showISBNx{978-1-939133-28-1}
\urldef\tempurl%
\url{https://www.usenix.org/conference/osdi22/presentation/zheng-lianmin}
\showURL{%
\tempurl}


\end{thebibliography}




\end{document}